\pdfoutput=1

\documentclass[11pt]{article}

\usepackage[final]{coling}

\usepackage{times}
\usepackage{latexsym}

\usepackage[T1]{fontenc}

\usepackage[utf8]{inputenc}

\usepackage{microtype}

\usepackage{inconsolata}

\usepackage{graphicx}
\usepackage{hyperref}
\usepackage{multirow}
\usepackage{booktabs}
\usepackage{algorithm,algcompatible,amsmath}
\usepackage{mathtools}
\usepackage{tabularx}

\def\FMT{\texttt{FaMSS}}
\newcommand{\methodFull}{Fact-aware Multilingual Selective Synergy}

\title{Selected Languages are All You Need for Cross-lingual Truthfulness Transfer}

\author{Weihao Liu\thanks{Work done during an internship at Microsoft.},\quad Ning Wu,\quad Wenbiao Ding, \\ \bf{Shining Liang,\quad Ming Gong,\quad Dongmei Zhang} \\
Microsoft STC Asia, Beijing \\ 
\texttt{liuweihao2022@outlook.com}\\
\texttt{\{wuning, wenbiaoding, shiningliang, migon, dongmeiz\}@microsoft.com} 
}
\begin{document}
\maketitle
\begin{abstract}

Truthfulness stands out as an essential challenge for Large Language Models~(LLMs). Although many works have developed various ways for truthfulness enhancement, they seldom focus on truthfulness in multilingual scenarios. Meanwhile, contemporary multilingual aligning technologies struggle to balance numerous languages and often exhibit serious truthfulness gaps across different languages, especially those that differ greatly from English. In our work, we extend truthfulness evaluation to multilingual contexts and propose a practical method for cross-lingual truthfulness transfer called \textit{\methodFull}~(\FMT). \FMT~is able to select an optimal subset of all tested languages by language bias and transfer contributions, and then employ translation instruction tuning for cross-lingual truthfulness transfer. Experimental results demonstrate that our approach can effectively reduce the multilingual representation disparity and boost cross-lingual truthfulness transfer of LLMs.~\footnote{\href{https://github.com/NeosKnight233/FaMSS}{https://github.com/NeosKnight233/FaMSS}}


\end{abstract}
\section{Introduction}

Large language models~(LLMs) have strong ability in generating human-level text in many domains~\cite{NEURIPS2020_1457c0d6, jiang2023mistral, touvron2023llama}. However, almost all LLMs face the issue of generating hallucinated responses~\cite{zhang2023sirens}, and demand methods to enhance their truthfulness. Although many methods for truthfulness enhancement in English have been proposed~\cite{tonmoy2024comprehensive, wang2023survey}, including designing new decoding strategies~\cite{shi2023trusting, chuang2023dola}, synthesizing high quality training data~\cite{tian2023fine} and representation editing~\cite{zhang2024truthx}, there is little attention to truthfulness in multilingual scenarios. Contemporary LLMs generally strive to achieve powerful multilingual capabilities, which promotes their applications worldwide~\cite{team2023gemini, llama3modelcard}. There, exploring the evaluation of truthfulness in multilingual contexts and cross-lingual truthfulness transfer holds significant importance.


\begin{figure}[tp]
    \centering
    \includegraphics[width= 0.48\textwidth]{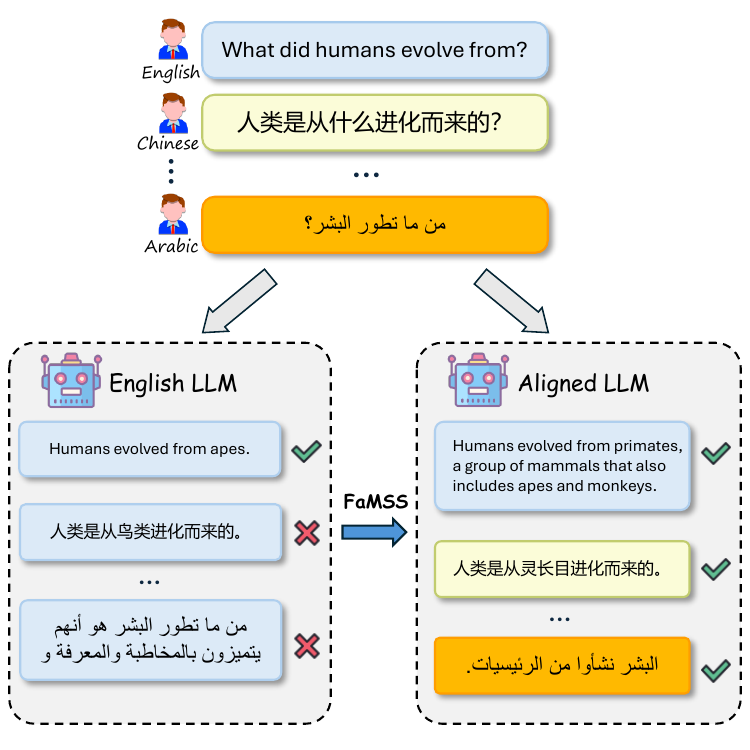}
    \caption{\FMT~is able to align the multilingual capabilities of LLMs, making them more truthful in answering multilingual questions.}
    \label{fig:fmt}
\end{figure}

To fill this gap of multilingual truthfulness, we construct a benchmark tailored for truthfulness evaluation in multilingual scenarios. Furthermore, we make preliminary explorations into methods aimed at cross-lingual truthfulness transfer of LLMs, with a focus on aligning the internal representations of factual descriptions. The most commonly used technique for building multilingual large models~(MLLMs) is cross-lingual instruction tuning~\cite{kulshreshtha2020cross, muennighoff2023crosslingual}, which usually requires synthesizing multilingual parallel corpus for various down-streaming tasks~\cite{alpaca, li2024eliciting} and struggles to overcome the alignment difficulties caused by introducing too many languages. In this work, we propose \textit{\methodFull}~(\FMT), a practical approach for multilingual alignment that relies solely on task-agnostic bilingual data. Meanwhile, \FMT~can achieve significant cross-lingual truthfulness transfer with selected optimal subset of tested languages by language bias and transfer contributions. As illustrated in Figure \ref{fig:fmt}, models utilizing \FMT~demonstrate higher truthfulness in multilingual question-answering tasks.




In summary, our contributions are as follows:
\begin{itemize}
    \item We construct \textbf{MTruthfulQA}, a novel benchmark designed to evaluate the truthfulness of LLMs in multilingual scenarios. This benchmark encompasses nine languages, each containing the same set of questions to ensure equitable evaluation of multilingual capabilities.
    \item We introduce a practical method for cross-lingual truthfulness transfer called \FMT. Through a data selection process prior to fine-tuning, \FMT~can efficiently and significantly boost truthfulness of LLMs across multiple languages.
    \item We systematically investigate how \texttt{FaMSS} facilitates the transfer of truthfulness across multiple languages. Based on our findings, we conclude that simply mixing training data from many different languages, which may interfere with each other, is not the most effective approach.
\end{itemize}

\section{Related Work}

\subsection{Truthfulness Evaluation Benchmark}


For question-answering tasks, TruthfulQA~\cite{lin2022truthfulqa} provides 817 challenging questions including 38 topics. FreshQA~\cite{vu2023freshllms} is a dataset aimed at testing LLMs on up-to-date world knowledge. For long-form text generation, FActScore ~\cite{min2023factscore} evaluates the factuality of LLMs by breaking down long texts into atomic claims. However, these classic benchmarks for truthfulness evaluation are all in English. In our work, we take a step towards evaluating and enhancing the truthfulness of LLMs in a multilingual setting. 


\subsection{Multilingual Ability Evaluation}

Previous works have provided several datasets and metrics for multilingual fact verification~\cite{gupta2021x, qi2023cross}, multilingual summarization~\cite{aharoni2022mface, qiu2023detecting} and open-book cross-lingual question-answering~\cite{mrtydi, lewis2020mlqa} tasks. With the development of LLMs, many more flexible and comprehensive evaluation datasets have been released. Meanwhile, many evaluation tasks have been expanded to multilingual settings. For instance, MGSM~\cite{shi2022language} evolved from GSM8K~\cite{cobbe2021training} provides mathematical ability evaluation in multilingual contexts. SeaEval~\cite{wang2023seaeval} hand-crafted high-quality Cross-MMLU and Cross-LogiQA datasets to test general knowledge question answering and reasoning capabilities of LLMs in multilingual scenarios. In addition to these, numerous studies have employed translation tools to construct relevant datasets or devise various metrics for MLLMs~\cite{lai2023okapi, lin2024crossin, chen2023breaking, shafayat2024multi, li-etal-2024-land}
. 



\subsection{Handling Numerous Languages}
Most works leverage cross-lingual pretraining and fine-tuning on different tasks to achieve multilingual capabilities in their models~\cite{qin2024multilingual, lample2019cross, huang-etal-2019-unicoder, devlin2019bert, xue2021mt5, muennighoff2023crosslingual, zhu2023extrapolating}. However, the difficulty in learning multiple languages increases significantly with the number of languages.
\citet{conneau2020unsupervised} discussed the challenges of scaling MLLMs to more languages, such as the \textit{curse of multilinguality}, where increasing the number of languages can dilute the model's capacity and potentially decrease overall performance. Therefore, some works mainly focus on training with high-resource languages\cite{yang2022hlt} or mining specific language pairs~\cite{fan2021beyond, lin2023mplm}. Other approaches propose clustering target languages into groups and training a single model for each cluster~\cite{tan2019multilingual, fan2021discovering}.

\section{Multilingual Truthfulness Benchmark}
\label{sec:benchmark}


\subsection{Data Source}

Since we mainly focus on the truthfulness~\cite{sun2024trustllm} of LLMs rather than their faithfulness~\cite{es2023ragas, maynez-etal-2020-faithfulness}, we prefer to consider context-free evaluation tasks. We employ TruthfulQA~\cite{lin2022truthfulqa}, a widely used dataset for evaluating the truthfulness of models in a question-answering format, as our foundation data. Compared with other QA datasets, TruthfulQA places greater emphasis on the ability to avoid false beliefs or misconceptions rather than complex reasoning.


\subsection{Multilingualism Formulation} 

To build a multilingual benchmark, we first select a few target languages from a diverse linguistic pool. Our evaluation system incorporates nine different languages: \textit{English~(En)}, \textit{French~(Fr)}, \textit{German~(De)}, \textit{Spanish~(Es)}, \textit{Chinese~(Zh)}, \textit{Japanese~(Ja)}, \textit{Russian~(Ru)}, \textit{Thai~(Th)} and \textit{Arabic~(Ar)}. For the questions and answers in original dataset, we translate them into the target languages through prompting \textit{GPT-4}~\footnote{Unless otherwise specified, we use GPT-4-1106 in this paper}. The prompt template we used is shown in Appendix~\ref{app:mtfqa_trans}.

Considering the instability of LLM-based translation, we also recruited native speakers to validate the translations. Each translation is reviewed by three native speakers, who vote on its correctness. If at least two of the three agree, the translation is considered accurate. The average accuracy of the translation is 96.4\% including both the questions and corresponding answers. We further manually checked and rectified these translation errors.




\subsection{Evaluation Metrics Migration}

We migrate and improve the evaluation metrics used in TruthfulQA ~\cite{lin2022truthfulqa}. For multi-choice QA, we adopt three likelihood scores $MC1$, $MC2$ and $MC3$, which is similar to~\citet{chuang2023dola}. Detailed calculations of metrics are reported in Appendix~\ref{app:metrics}.

For open-ended generation, we replace the evaluation model in the original pipeline with a stronger open source model Mistral-7b~\footnote{\href{https://huggingface.co/mistralai/Mistral-7B-Instruct-v0.2}{Mistral-7b-Instruct-v0.2}}~\cite{jiang2023mistral}. We then utilize a similar translation pipeline to build the multilingual classification dataset and fine-tune two models for truthfulness and informativeness evaluation, which we refer to as "MM-Judge ~(Multilingual-Mistral-Judge)". An acceptable answer should be both truthful and informative (e.g. "I have no comment." is truthful but lacks valuable information). These two scores to judge whether the answer is truthful and informative are denoted as True~(\%) and Info~(\%) respectively. The product True*Info~(\%) serves as a more comprehensive and reasonable metric. The reliability of "MM-Judge" is discussed and verified in Appendix~\ref{app:mm-judge}.

\section{\methodFull}

To balance and enhance truthfulness in different languages, we propose \methodFull~(\FMT), which can effectively fine-tune models for multilingualism with fact-aware multilingual data in a selective languages synergy manner.


\subsection{Fact-aware Multilingual Data}


Our primary training data is derived from a variety of parallel corpus. Although there are many kinds of parallel corpora used in machine translation, including WikiMatrix~\cite{schwenk2021wikimatrix}, UNPC~\cite{ziemski2016united}, Tatoeba~\cite{tiedemann2020tatoeba}, etc., we consider not using too much common machine translation corpora. These MT corpora mainly consist of simple sentences in each language with limited factual descriptions or only focus on specific areas, which does not contribute much to improving the truthfulness of models. Consequently, we employ fact-aware multilingual data, which includes rich bilingual factual descriptions alongside high-quality parallel corpora. In general, we include three different types of data: 

\paragraph{Factuality Translation Data} Similar to MLQA~\cite{lewis2020mlqa}, we first perform parallel sentence mining over Wikipedia articles through the LASER toolkit\footnote{\href{https://github.com/facebookresearch/LASER}{https://github.com/facebookresearch/LASER}}~\cite{artetxe2019massively} with some specific topics, including \textit{history}, \textit{biography}, \textit{geography}, \textit{science} and \textit{cultural}. For each parallel sentence, we extract a broader context~(e.g. a paragraph) that provides additional factual details related to the sentence. This encourages the factual integrity and depth of the content. Given the challenges of exploring fully aligned context across nine languages, we also maintain a 4-way aligned approach, which means that a factual description appears in at least four different languages~(always including English). 

\paragraph{Common Translation Data} In our study, we find that a certain amount of common translation data is still beneficial for multilingual alignment. FLORES-200~\cite{costa2022no} provides high-quality translation data between more than 200 languages, making it a suitable choice for our purposes.

\paragraph{Pretraining Data} \citet{dou2024loramoe} find that excessive amount of instruction data during supervised fine-tuning~(SFT) can degrade the world knowledge stored in LLMs, leading to performance drop in knowledge tasks. Our experiments also support this conclusion. Thus, besides translation data, we incorporate extra English pretraining data from Wikipedia to alleviate the world knowledge forgetting. Note that we usually have about 10\% of the training data as pretraining data.

We have data of all nine languages mentioned in section \ref{sec:benchmark}, which means a massive number from possible translation directions~($9\times8=72$) for a single example. Therefore, similar to \citet{zhu2023extrapolating}, we follow the most widely used English-centric way and further adopt the best translation direction, which means that we only put the non-English text on the target side and the English text on the source side. The details of our collected training data are presented in Appendix~\ref{app:exampledata}.

\subsection{Selective Languages Synergy}

Incorporating a large number of languages complicates the alignment process, which is known as the \textit{curse of multilinguality}~\cite{conneau2020unsupervised}. Therefore, we aim to select a core subset of languages, which helps align all nine languages in a collaborative and diffusive manner. This idea also coincides with Pareto principle~\cite{sanders1987pareto}, which indicates that the most important factors in a system are only about 20\% but affect 80\% of the results. In the selection process, we first propose a \textbf{language bias probe} to help cluster languages into several groups and then select core languages from different groups.
\begin{figure}[h]
    \centering
    \includegraphics[width= 0.48 \textwidth]{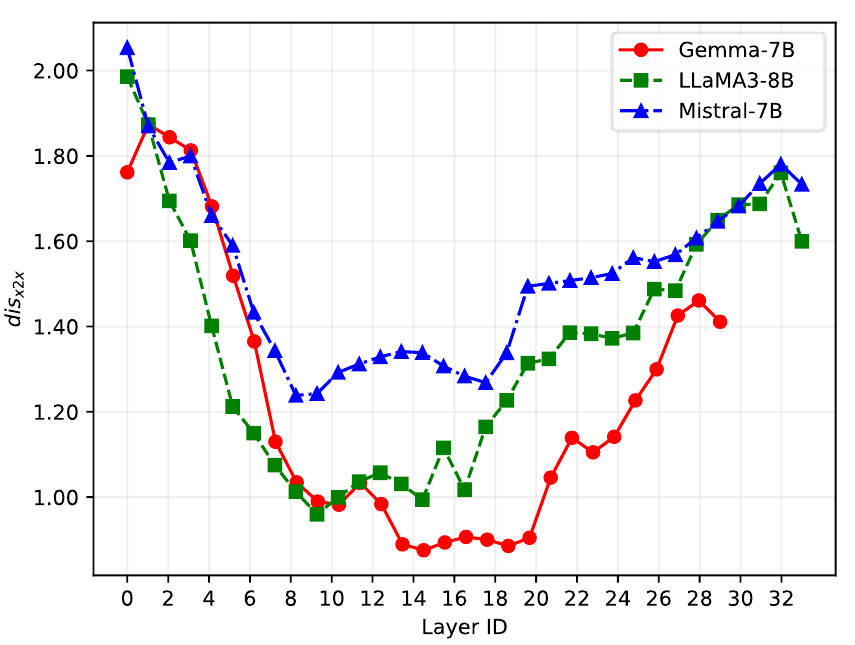}
    \caption{Mean bias between languages. $dis_{x2x}$ represents the average distance value of all language pairs in one layer.}
    \label{fig:distance}
\end{figure}

\begin{algorithm}[h]
    \renewcommand{\algorithmicrequire}{\textbf{Input:}}
    \renewcommand{\algorithmicensure}{\textbf{Output:}}
    \caption{\textbf{Language Bias Probe}}
    \label{algo:probe}
    \begin{algorithmic}[1]
        \REQUIRE Model $\mathcal{M}$, parallel corpus $\mathcal{C}$, language set $\mathcal{L}$.
        \ENSURE Bias $\mathcal{D}$ between any language pair in $\mathcal{L}$ of each layer in $\mathcal{M}$.
        \STATE $\mathcal{L}$ $\gets$ \{"English", "Chinese", "Spanish", $\dots$\}
        \STATE $N \gets $ number of decoder layers in $\mathcal{M}$ 
        \STATE $M \gets $ number of samples in $\mathcal{C}$
        
        
        \FOR{$i = 0$ \textbf{to} $N$}
            \STATE $\mathcal{H}[i] \gets$ mean sentence hidden states calculated from $\mathcal{C}$ at layer $i$
            \STATE $\mathcal{H}[i] \gets$ Standardization($\mathcal{H}[i]$)
            
            \FOR{$(l_j, l_k)$ \textbf{in} $(\mathcal{L},\mathcal{L})$}
                
                \FOR{$m = 0$ \textbf{to} $M-1$}
                    \STATE $d \gets \Vert (\mathcal{H}[i][l_j][m]-\mathcal{H}[i][l_k][m] )\Vert_2^2$
                    \STATE $\mathcal{D}[i][l_j][k] \gets \mathcal{D}[i][l_j][k] + d$ 
                \ENDFOR
                \STATE $\mathcal{D}[i][l_j][l_k] \gets \mathcal{D}[i][l_j][l_k]/M$
            \ENDFOR
        \ENDFOR
        \STATE Return $\mathcal{D}$
    \end{algorithmic}
\end{algorithm}

\subsubsection{Language Bias Probe}

Different languages have different characteristics, and some of them may share  representation similarities or show big differences, which can be used to guide the selection of training data. We propose a language bias probe to estimate the representation bias of models between languages. Inspired by the findings of \citet{ju2024large} that different layers in LLMs may encode different contextual information, we hypothesize that models exhibit minimal mean language bias in the layer where they encode richest semantic representations. For convenience, we will refer to the layer encoding the most language-agnostic semantic information as the semantic layer.

Formally, we define the bias between each language pair as their average representation distance on the probe corpora $\mathcal{C}$, denoted as $\mathcal{D}$. The algorithm to calculate $\mathcal{D}$ is described in Algorithm \ref{algo:probe}, where $\mathcal{C}$ comes from FLORES-200~\cite{costa2022no} in our work. As shown in Figure \ref{fig:distance}, the mean bias of languages in three popular foundation models initially decreases and then increases as the index of decoder layers increases. Thus, we can infer that the lowest point on each curve indicates the semantic layer of each foundation model~(e.g., layer 14 may be the semantic layer of \textit{Gemma-7B}). We will leverage this information to analyze the transfer contributions of specific languages and optimize the combination of training data. 

\subsubsection{Probing to Select}
\label{subsec:probing}
Suppose that we have $M$ languages in our training corpora $\mathcal{C}$ and evaluation set denoted as $\mathcal{L} = \{l_1,l_2,\dots,l_M\}$. Our goal is to select an optimal language set $s_o$ from $\mathcal{L}$, which can boost the alignment between English and other non-English languages as effectively as possible. 

We take two major steps to form $s_o$: 1) group all languages into several clusters and 2) select a core language from each cluster as one element in $s_o$. For step 1, we cluster languages by language bias. For step 2, we define a transfer contribution $\mathcal{TC}$ for each language:
\begin{equation}
    \mathcal{TC}_l = \sum_{l'\in \mathcal{L}} \mathcal{D}_s[\text{"English"}][l']-\mathcal{D}^l_s[\text{"English"}][l'] \nonumber
\end{equation}

\begin{algorithm}[h]
    \renewcommand{\algorithmicrequire}{\textbf{Input:}}
    \renewcommand{\algorithmicensure}{\textbf{Output:}}
    \caption{Optimal Language Set Selection}
    \label{algo:lan_select}
    \begin{algorithmic}[1]
        \REQUIRE Language set $\mathcal{L}$, language bias matrix $\mathcal{D}_s$ of the semantic layer, transfer contribution of each language $\mathcal{TC}$, maximum number of languages $m$, distance threshold $d$.
        \ENSURE Optimal language set $s_o$
        \STATE $\mathcal{S} \gets \{\ \{l\}\ |\ l\in \mathcal{L} \}$
        \WHILE {$|\mathcal{S}|>m$}
            \STATE $\mathcal{S}' \gets \mathcal{S}$
            
            \FOR{each $s \in \mathcal{S}$}
                \STATE $s' \gets$ nearest\_set($s$, $\mathcal{S}$, $\mathcal{D}_s$, $d$)
                \IF{$s'$ exists}
                    \STATE $\mathcal{S}'\gets (\mathcal{S'} \textbackslash \{s, s'\}) \cup \{s\cup s'\}$
                \ENDIF
            \ENDFOR
            
            \IF{$|\mathcal{S}|=|\mathcal{S}'|$}
                \STATE break
            \ENDIF
            \STATE $S \gets S'$
        \ENDWHILE
        \STATE $s_o \gets$ \{element\_with\_max\_$\mathcal{TC}$($s$)\ |\ $s \in \mathcal{S}$\}
        \STATE Return $s_o$
    \end{algorithmic}
\end{algorithm}

where $\mathcal{D}_s$, $\mathcal{D}^l_s$ means the distance matrix of the semantic layer before and after fine-tuning on language $l$. The language with largest $\mathcal{TC}$ in corresponding cluster is selected as the core language of that cluster. Algorithm~\ref{algo:lan_select} illustrates the algorithm to select optimal language set $s_o$, where the role of the parameters $m$ and $d$ is to control the number of languages involved in the training stage and to avoid merging languages with too large differences into the same set. For the nearest set, we define the distance between two sets $s$ and $s'$ as the minimum language bias among all language pairs $(l_1, l_2)$ in $s\times s'$. For more details about the selecting process, please refer to Appendix~\ref{app:detail_selection}.







\subsubsection{Translation Instruction Tuning}

We adopt the instruction tuning~\cite{zhang2023instruction} pipeline leveraging constructed bilingual data. We use the translation instruction template in Figure~\ref{fig:training_format} of Appendix~\ref{app:training_format}, and fine-tuning models by next-token prediction with cross-entropy loss. 

\begin{table*}[htb]
\centering
\resizebox{0.9 \textwidth}{!}{
\begin{tabular}{l|cccccccccc}
\toprule

\multirow{2}{*}{\textbf{Models}} & \multicolumn{10}{c}{\textbf{True*Info(\%)}}                                                                                                                       \\ \cline{2-11} 
                                 & \textbf{En}   & \textbf{De}   & \textbf{Fr}   & \textbf{Es}   & \textbf{Ru}   & \textbf{Zh}   & \textbf{Ja}   & \textbf{Th}   & \textbf{Ar}   & \textbf{Avg.} \\ \hline
\textit{GPT-4}~\cite{achiam2023gpt}                           & 69.8          & 71.4          & 72.5          & 67.9          & 68.9          & 68.9          & 68.1          & 56.5          & 68.5          & 68.1          \\
\textit{Bloomz-7B1-mt}~\cite{muennighoff2023crosslingual}                    & 18.7          & 21.9          & 20.7          & 13.2          & 16.0          & 26.8          & 16.4          & 13.5          & 14.2          & 17.9          \\
\textit{LLaMA-3-8B-Instruct}~\cite{llama3modelcard}              & 56.7          & 30.7          & 43.7          & 29.1          & 37.3          & 34.4          & 18.7          & 17.3          & 31.6          & 33.3          \\
\textit{Mistral-7B-Instruct-v0.2}~\cite{jiang2023mistral}         & 66.7          & 52.5          & 57.9          & 49.9          & 49.1          & 52.9          & 27.7          & 24.6          & 42.7          & 47.1          \\ 
\hline
\textbf{With \FMT}               &               &               &               &               &               &               &               &               &               &               \\ \hline
\textit{LLaMA-3-8B}~\cite{llama3modelcard}                       & 36.6          & 31.6          & 31.1          & 27.3          & 38.8          & 31.1          & 25.1          & 20.1          & 42.1          & 31.5          \\
\textit{LLaMA-3-8B} + \textbf{\FMT}                & \textbf{44.9} & \textbf{34.9} & \textbf{34.5} & \textbf{37.3} & \textbf{42.4} & \textbf{36.2} & \textbf{39.9} & \textbf{24.7} & 41.9          & \textbf{37.4} \\ \hline
\textit{Mistral-7B-v0.1}~\cite{jiang2023mistral}          & 36.0          & 35.9          & 28.8          & 29.3          & 34.0          & 35.9          & 19.5          & 15.4          & 22.2          & 28.5          \\
\textit{Mistral-7B-v0.1} + \textbf{\FMT}                  & \textbf{39.6}          & \textbf{41.2}          & \textbf{31.4}          & \textbf{29.3}          & 33.0          & \textbf{40.0}          & \textbf{20.5}          & 14.5          & \textbf{29.6}          & \textbf{31.0}          \\ \hline
\textit{Gemma-7B}~\cite{team2024gemma}                      & 25.7          & 19.7          & 17.7          & 17.1          & 21.5          & 27.8          & 34.6          & 15.8          & 19.7          & 22.2          \\
\textit{Gemma-7B} + \textbf{\FMT}               & \textbf{37.8} & \textbf{45.9} & \textbf{22.2} & \textbf{23.6} & \textbf{30.8} & \textbf{41.7} & \textbf{40.1} & \textbf{19.7} & \textbf{25.2} & \textbf{31.9} \\
\bottomrule
\end{tabular}}
\caption{Experimental results on MTruthfulQA dataset under open-ended generation setting. Presented metric is True*Info~(\%) score, representing ability to generate truthful and informative contents. Full results can be found in Appendix \ref{app:result}.}
\label{tab:TI}

\end{table*}

\begin{table*}[htb]
\centering
\resizebox{0.8 \textwidth}{!}{
\begin{tabular}{l|cccccccccc}
\toprule

\multirow{2}{*}{\textbf{Models}} & \multicolumn{10}{c}{\textbf{MC1~(\%)}}                                                                                                           \\ \cline{2-11} 
                                 & \textbf{En} & \textbf{De} & \textbf{Fr} & \textbf{Es} & \textbf{Ru} & \textbf{Zh} & \textbf{Ja} & \textbf{Th} & \textbf{Ar} & \textbf{Avg.} \\ \hline
\textit{Bloomz-7B1-mt}                    & 26.7        & 27.1        & 27.3        & 24.4        & 29.0        & 26.2        & 29.7        & 26.2        & 27.7        & 27.1          \\
\textit{LLaMA-3-8B-Instruct}              & 40.6        & 36.4        & 40.5        & 37.5        & 35.7        & 37.9        & 36.0        & 28.5        & 35.7        & 36.5          \\
\textit{Mistral-7B-Instruct-v0.2}         & 55.4        & 49.7        & 52.8        & 47.4        & 44.7        & 42.2        & 40.3        & 29.7        & 37.5        & 44.4          \\ 
\hline
\textbf{With \FMT}               &             &             &             &             &             &             &             &             &             &               \\ \hline
\textit{LLaMA-3-8B}                       & 32.4        & 31.5        & 33.5        & 30.5        & 31.6        & 28.9        & 34.6        & 28.3        & 29.9        & 31.2          \\
\textit{LLaMA-3-8B} + \textbf{\FMT}                & \textbf{35.0}        & \textbf{31.7}        & 32.7        & \textbf{34.5}        & \textbf{32.8}        & \textbf{30.2}        & \textbf{35.5}        & 26.9        & \textbf{30.1}        & \textbf{32.2}          \\ \hline
\textit{Mistral-7B-v0.1}                  & 30.6        & 33.5        & 33.0        & 29.7        & 33.0        & 32.1        & 32.9        & 28.2        & 31.3        & 31.6          \\
\textit{Mistral-7B-v0.1} + \textbf{\FMT}  & \textbf{32.8}        & \textbf{33.7}        & 32.8        & \textbf{31.6}        & \textbf{33.4}        & \textbf{32.9}        & 32.7        & 27.7        & \textbf{31.8}        & \textbf{32.1}          \\ \hline
\textit{Gemma-7B}                      & 33.4        & 32.4        & 33.2        & 33.4        & 30.5        & 29.9        & 33.0        & 27.7        & 31.7        & 31.7          \\
\textit{Gemma-7B }+ \textbf{\FMT}               & \textbf{37.3}        & \textbf{34.8}        & \textbf{36.2}        & \textbf{33.9}        & \textbf{31.3}        & \textbf{30.4}        & \textbf{33.4}        & \textbf{28.9}        & 31.6        & \textbf{33.1} \\
\bottomrule
\end{tabular}}
\caption{Experimental results on MTruthfulQA dataset under multi-choice QA setting. Presented metrics are $MC1$ scores mentioned in section \ref{sec:benchmark}. Full results can be found in Appendix \ref{app:result}.}
\label{tab:MC}
\end{table*}

\section{Experiments}

\subsection{Setup}


In our experiments, we utilize three representative open source foundation LLMs as our base models: \textit{LLaMA-3-8B}~\cite{llama3modelcard}, \textit{Mistral-7B-v0.1}~\cite{jiang2023mistral} and \textit{Gemma-7B}~\cite{team2024gemma}. We employ full parameters fine-tuning over selected subset of fact-aware multilingual data. Details about our training are placed in Appendix \ref{app:translate_detail}.


For evaluation benchmarks, we first evaluate our models on MTruthfulQA we construct in both multi-choice QA and open-ended generation settings. We also do evaluations on Cross-MMLU~\cite{wang2023seaeval}, which contains 150 high-quality human-annotated questions for common knowledge testing across seven languages: \textit{English}, \textit{Chinese}, \textit{Indonesian}, \textit{Spanish}, \textit{Vietnamese}, \textit{Malay}, and \textit{Filipino}.

\subsection{Main Results}


\begin{figure*}[htbp]
    \centering
    \includegraphics[width= 0.88 \textwidth]{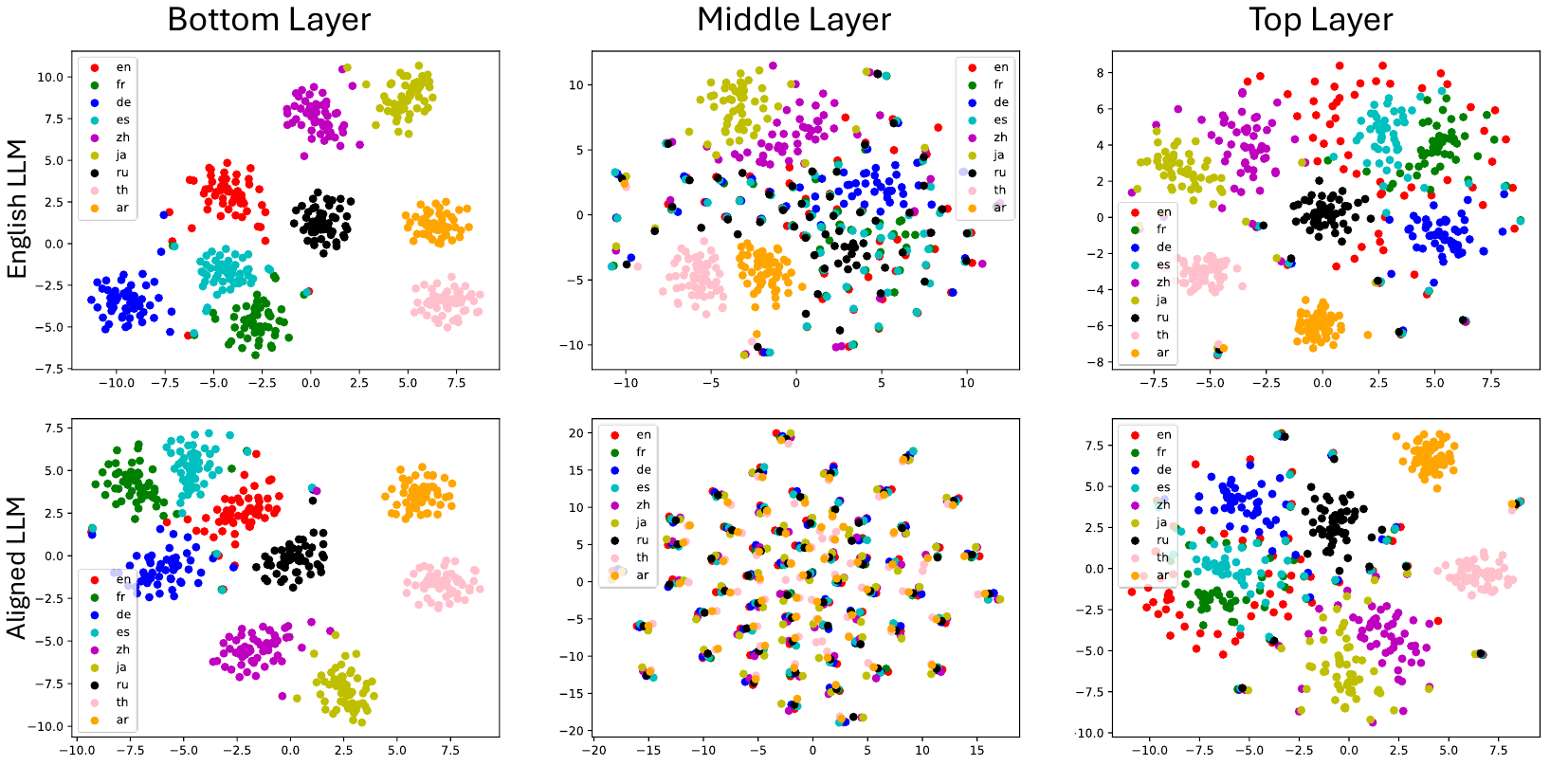}
    \caption{Visualization results in the representation space of LLMs on FLORES-200 before and after utilizing~\FMT. Our method successfully enhances the multilingual representation alignment in the middle layers, where models encode more semantic-level information. In contrast, the top and bottom layers contain more syntactic information, resulting in little overlap in representations across different languages.}
    \label{fig:tsne}
\end{figure*}

Table \ref{tab:TI} and Table \ref{tab:MC} present the results of models on MTruthfulQA. Following the standard setting from~\citet{lin2022truthfulqa} and~\citet{zhang2024truthx}, all experiments are performed in a few-shot prompting setting. We analyze the performance of all models in depth and draw some conclusions. 
\paragraph{MTruthfulQA reveals the ability of LLMs' truth-telling and truth-recognizing in multilingual scenarios.} First, in open-ended generation evaluation, strong baselines such as \textit{GPT-4}~\cite{achiam2023gpt} and \textit{Mistral-7B-Instruct-v0.2}~\cite{jiang2023mistral} get much higher scores compared with other models, showing more powerful truth-telling ability across massive languages. Similar results are observed for truth-recognizing in multi-choice setting. Furthermore, we can obviously see models exhibit different truthfulness levels across different languages, usually higher in English or other languages similar to English. This is reasonable given that LLMs are trained with more data in these languages. Additionally, the gap between different language pairs is not uniform in the two evaluation settings. We can see larger gap in True*Info~(\%) scores compared with \textit{MC} scores, indicating that open-ended generation is a more challenging task and better distinguishes the multilingual capabilities of models.

\paragraph{\FMT~is convenient and effective.} Although we just train our models with a majority of translation instruction data, our method achieves great enhancement of truthfulness in tested foundation models. We get \textbf{+5.9\%} of improvement in True*Info~(\%) score on \textit{LLaMA-3-8B}, \textbf{+2.5\%} on \textit{Mistral-7B-v0.1} and \textbf{+9.7\%} on \textit{Gemma-7B}. Meanwhile, the result of multi-choice QA on MTruthfulQA is also averagely positive. Additionally, models fine-tuned with \FMT~show stronger multilingual capabilities in common knowledge benchmark. As shown in Table \ref{tab:crossmmlu}, our method is significantly effective on Cross-MMLU, raising accuracy of prediction by \textbf{+5.2\%} on \textit{LLaMA-3-8B}, \textbf{+1.4\%} on \textit{Mistral-7B-v0.1} and \textbf{+5.6\%} on \textit{Gemma-7B}. A notable phenomenon is that the performance improvement of \textit{Mistral-7B-v0.1} is relatively small. This may be due to the relatively limited vocabulary size of it compared with other models, which restricts the efficiency of multilingual alignment and highlights the importance of a large vocabulary for MLLMs. In general, these positive results highlight the effectiveness of \FMT, showing that fact-aware multilingual data can be conveniently used to enhance multilingual truthfulness.

\begin{table}[htb]
\centering
\resizebox{0.43 \textwidth}{!}{
\begin{tabular}{l|ccc}
\toprule
\textbf{Model}           & \textbf{Acc}  & \textbf{Consis} & \textbf{AC3}  \\ \hline
\textit{GPT-4}                    & 85.0          & 80.0            & 83.0          \\
\textit{LLaMA3-8B-Instruct}       & 57.0          & 48.0            & 52.0          \\
\textit{Mistral-7B-Instruct-v0.2} & 49.0          & 30.0            & 37.0          \\ 
\hline
\textbf{With \FMT}        &               &                 &               \\ \hline
\textit{LLaMA-3-8B}               & 46.8          & 28.6            & 35.5          \\
\textit{LLaMA-3-8B} + \textbf{\FMT}         & \textbf{52.0} & \textbf{40.8}   & \textbf{45.7} \\ \hline
\textit{Mistral-7B-v0.1} & 51.2          & 45.1            & 48.0          \\ 
\textit{Mistral-7B-v0.1} + \textbf{\FMT} & \textbf{52.6}          & \textbf{46.3}            & \textbf{49.2 }         \\ \hline
\textit{Gemma-7B}              & 48.4          & 54.2            & 51.1          \\
\textit{Gemma-7B} + \textbf{\FMT}        & \textbf{54.0} & 52.0            & \textbf{53.0} \\
\bottomrule
\end{tabular}}
\caption{Experimental results on Cross-MMLU dataset.}
\label{tab:crossmmlu}
\end{table}

\paragraph{Trained and not directly trained languages can both benefit from \FMT.} We investigate the difference in performance change between trained languages and untrained languages. For example, when we fine-tuned \textit{Gemma-7B} on language set $\mathcal{S}=\{$German, Chinese, Arabic$\}$, our tuning method get higher True*Info~(\%) scores in these languages in $\mathcal{S}$ compared to the base model, which can be regarded as these languages "learn" from English during \FMT. On the other hand, languages not in $\mathcal{S}$ also benefit from the alignment procedure between English and $\mathcal{S}$, showing relatively smaller improvement on True*Info~(\%) score. Moreover, the performance gain on Cross-MMLU further support this conclusion, as there are very few languages that are both included in Cross-MMLU and $\mathcal{S}$~(i.e., just Chinese since English is the source language). To get a deeper understanding of the enhancement in semantic alignment between languages brought by \FMT, we also present a visual comparison through the t-SNE~\footnote{\href{https://scikit-learn.org/stable/}{https://scikit-learn.org/stable/}} algorithm in Figure~\ref{fig:tsne}.





\section{Ablation and Further Analysis}


To further explore the influence of \FMT~across languages and explain the data allocation schema used in our experiments, we conduct ablation studies and additional analyses. The reported results are obtained on \textit{Gemma-7B}, but similar conclusions were found across other models as well.


\subsection{Data Allocation of Languages}
\label{sec:languagesallocation}

\paragraph{Alignment between multiple languages may not be performed synchronously.}

\begin{figure}[ht]
    \includegraphics[width= 0.48 \textwidth]{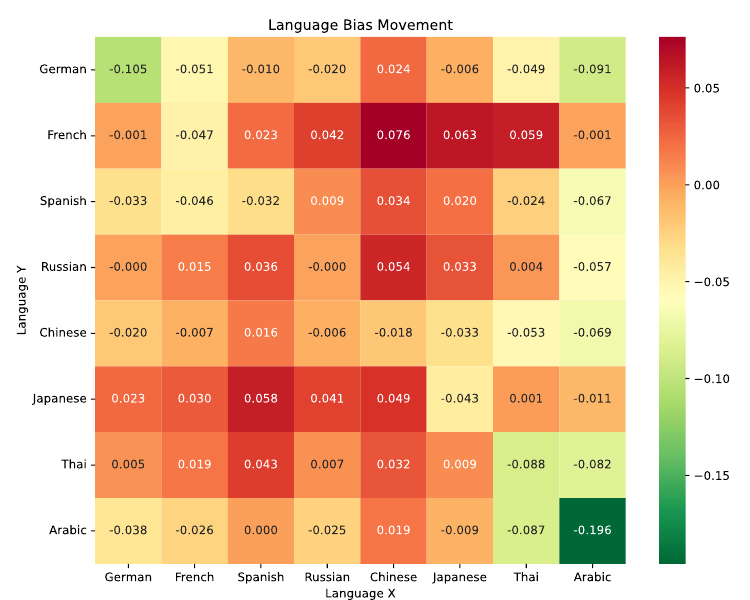}
    \caption{Language bias movement when fine-tuned on language Y and probed on language X. All displayed values are calculated on the 14th layer of \textit{Gemma-7B}. The value can also be formulated as $\mathcal{D}_s^Y[\text{"English"}][X]-\mathcal{D}_s[\text{"English"}][X]$.}
    \label{fig:biasmov}
\end{figure}

Ideally, we would combine data from all languages and hope for an aligning process between each individual language and English. However, the representation of English text in semantic space cannot simultaneously align with all other languages. In fact, languages do interfere with each other when trained together. To demonstrate this point, we separately fine-tuned the base model with data in each language and leveraged the language bias probe to measure the movement of language bias between English and other languages. As shown in Figure \ref{fig:biasmov}, the representation of a given language $X$ consistently becomes closer to English when fine-tuned on translation data in language $X$. However, for some languages, this can also cause a larger gap between English and another language $Y$. For instance, fine-tuning on translating English to French makes French closer to English but negatively impacts the alignment between English and Chinese.

\paragraph{Combination of a few languages is more robust.} 


\begin{figure}[htbp]
    \centering
    \includegraphics[width= 0.5\textwidth]{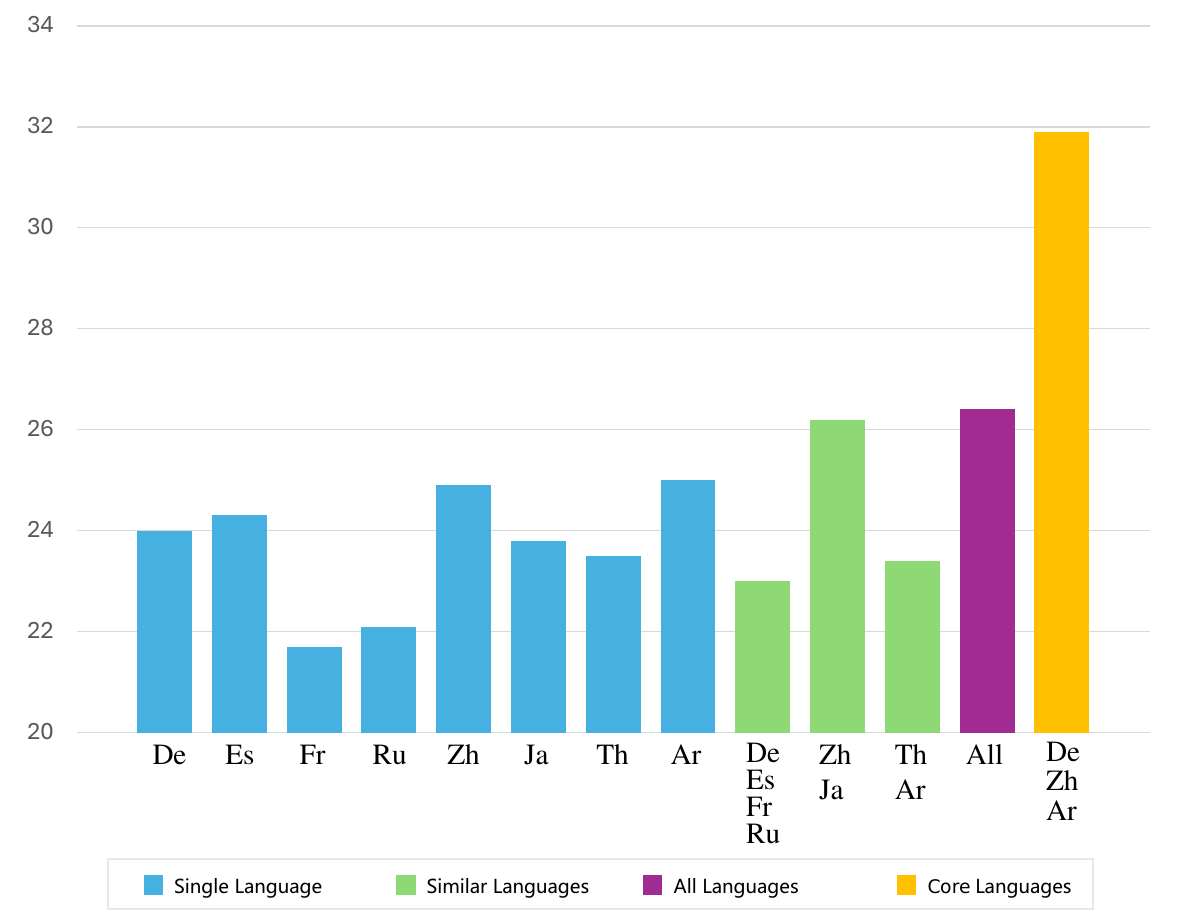}
    \caption{Performance on MTruthfulQA when fine-tuned with different language sets. }
    \label{fig:langset}
\end{figure}

One main advantage of our leveraged \FMT~method is that we can fine-tune models with just a subset of target languages while achieving more robust performance. To further confirm the rationality behind selective languages synergy, we compared the model's performance under different language selection settings. We report the True*Info~(\%) scores of training with different language sets on \textit{Gemma-7B} in Figure \ref{fig:langset}. The results show that simply mixing training data from similar languages may actually harm the model's performance, even if the total amount of training data increases. This indicates that training with data from different languages makes it more challenging for the model to improve its overall multilingual truthfulness. In contrast, selecting only three core languages instead of the whole language set enables the model to learn the intrinsic connections between multiple languages more efficiently, thereby achieving a higher level of multilingual proficiency. We also conducted ablation studies on the performance change when the language set grew larger with different $m$ and $d$ values defined in section~\ref{subsec:probing}, and the results in Table~\ref{tab:dif_md} in Appendix~\ref{app:result} also confirmed that increasing the number of languages could potentially impair the overall multilingual truthfulness. 

\subsection{Effectiveness of Mixture Data Types}



\begin{figure}[h]
    \centering
    \includegraphics[width= 0.5\textwidth]{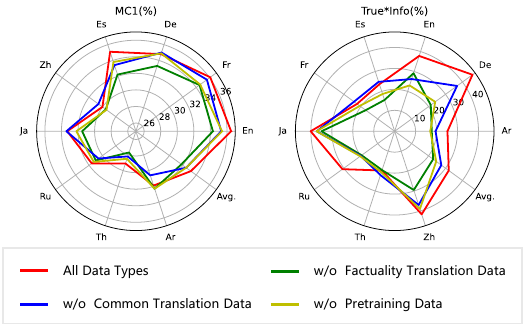}
    \caption{Multilingual performance of LLMs with different mixture of data types.}
    \label{fig:radar}
\end{figure}

Figure~\ref{fig:radar} displays the results on \textit{Gemma-7B} with different mixtures of data types. The optimal performance is achieved with a balanced integration of all three data types. The synergistic effect of a diverse data mixture may arise from the complementary nature of different data types~(red line). 

The omission of factuality data~(green line) leads to a noticeable struggle in enhancing the truthfulness of LLMs. This is primarily due to the model's reduced capacity to discern and generate factually accurate information. Meanwhile, the inclusion of the other two types of data further improves the truthfulness level across most languages.




\section{Conclusion}

In this paper, we construct a novel MTruthfulQA benchmark, which enables the evaluation of truthfulness in multilingual scenarios. Through our experiments, we draw the conclusion that simply mixing data of  languages together results in a low cross-lingual truthfulness transfer efficiency and harm the overall performance. In the contrast, our proposed method \FMT~successfully boosts cross-lingual truthfulness transfer with less well-selected data.





\section*{Limitations}

In our study, we primarily evaluate our models on closed-book QA tasks. It remains unclear whether our method is equally effective on extractive QA or other context-demanded tasks, which are more related to the faithfulness of LLMs. Regarding the data allocation of different languages, it is possible that incorporating a small proportion of not considered languages rather than entirely omitting them, might yield better results. Considering the numerous scenarios introduced by varying data proportions, we leave it as future research to explore the impact of diverse language ratios on the model's overall multilingual performance.


\section*{Ethical Considerations}


The translated multilingual data utilized in this work may not be perfectly aligned with the original source data and could contain some unreasonable descriptions. We have made every effort to mitigate these issues to the best of our ability. Additionally, the results in our language selection process might lead to misconceptions regarding the superiority or inferiority of certain languages; however, our work does not contain any bias against any language. Our work strictly adheres to the license and policies of released LLMs and publicly available datasets.


\bibliography{custom}

\begin{thebibliography}{59}
\providecommand{\natexlab}[1]{#1}

\bibitem[{Achiam et~al.(2023)Achiam, Adler, Agarwal, Ahmad, Akkaya, Aleman, Almeida, Altenschmidt, Altman, Anadkat et~al.}]{achiam2023gpt}
Josh Achiam, Steven Adler, Sandhini Agarwal, Lama Ahmad, Ilge Akkaya, Florencia~Leoni Aleman, Diogo Almeida, Janko Altenschmidt, Sam Altman, Shyamal Anadkat, et~al. 2023.
\newblock Gpt-4 technical report.
\newblock \emph{arXiv preprint arXiv:2303.08774}.

\bibitem[{Aharoni et~al.(2022)Aharoni, Narayan, Maynez, Herzig, Clark, and Lapata}]{aharoni2022mface}
Roee Aharoni, Shashi Narayan, Joshua Maynez, Jonathan Herzig, Elizabeth Clark, and Mirella Lapata. 2022.
\newblock mface: Multilingual summarization with factual consistency evaluation.
\newblock \emph{arXiv preprint arXiv:2212.10622}.

\bibitem[{AI@Meta(2024)}]{llama3modelcard}
AI@Meta. 2024.
\newblock \href {https://github.com/meta-llama/llama3/blob/main/MODEL_CARD.md} {Llama 3 model card}.

\bibitem[{Artetxe and Schwenk(2019)}]{artetxe2019massively}
Mikel Artetxe and Holger Schwenk. 2019.
\newblock Massively multilingual sentence embeddings for zero-shot cross-lingual transfer and beyond.
\newblock \emph{Transactions of the association for computational linguistics}, 7:597--610.

\bibitem[{Brown et~al.(2020)Brown, Mann, Ryder, Subbiah, Kaplan, Dhariwal, Neelakantan, Shyam, Sastry, Askell, Agarwal, Herbert-Voss, Krueger, Henighan, Child, Ramesh, Ziegler, Wu, Winter, Hesse, Chen, Sigler, Litwin, Gray, Chess, Clark, Berner, McCandlish, Radford, Sutskever, and Amodei}]{NEURIPS2020_1457c0d6}
Tom Brown, Benjamin Mann, Nick Ryder, Melanie Subbiah, Jared~D Kaplan, Prafulla Dhariwal, Arvind Neelakantan, Pranav Shyam, Girish Sastry, Amanda Askell, Sandhini Agarwal, Ariel Herbert-Voss, Gretchen Krueger, Tom Henighan, Rewon Child, Aditya Ramesh, Daniel Ziegler, Jeffrey Wu, Clemens Winter, Chris Hesse, Mark Chen, Eric Sigler, Mateusz Litwin, Scott Gray, Benjamin Chess, Jack Clark, Christopher Berner, Sam McCandlish, Alec Radford, Ilya Sutskever, and Dario Amodei. 2020.
\newblock \href {https://proceedings.neurips.cc/paper_files/paper/2020/file/1457c0d6bfcb4967418bfb8ac142f64a-Paper.pdf} {Language models are few-shot learners}.
\newblock In \emph{Advances in Neural Information Processing Systems}, volume~33, pages 1877--1901. Curran Associates, Inc.

\bibitem[{Chen et~al.(2023)Chen, Zheng, Wu, Shou, Gong, Song, Zhang, and Li}]{chen2023breaking}
Nuo Chen, Zinan Zheng, Ning Wu, Linjun Shou, Ming Gong, Yangqiu Song, Dongmei Zhang, and Jia Li. 2023.
\newblock Breaking language barriers in multilingual mathematical reasoning: Insights and observations.
\newblock \emph{arXiv preprint arXiv:2310.20246}.

\bibitem[{Chuang et~al.(2023)Chuang, Xie, Luo, Kim, Glass, and He}]{chuang2023dola}
Yung-Sung Chuang, Yujia Xie, Hongyin Luo, Yoon Kim, James Glass, and Pengcheng He. 2023.
\newblock Dola: Decoding by contrasting layers improves factuality in large language models.
\newblock \emph{arXiv preprint arXiv:2309.03883}.

\bibitem[{Cobbe et~al.(2021)Cobbe, Kosaraju, Bavarian, Chen, Jun, Kaiser, Plappert, Tworek, Hilton, Nakano et~al.}]{cobbe2021training}
Karl Cobbe, Vineet Kosaraju, Mohammad Bavarian, Mark Chen, Heewoo Jun, Lukasz Kaiser, Matthias Plappert, Jerry Tworek, Jacob Hilton, Reiichiro Nakano, et~al. 2021.
\newblock Training verifiers to solve math word problems.
\newblock \emph{arXiv preprint arXiv:2110.14168}.

\bibitem[{Conneau et~al.(2020)Conneau, Khandelwal, Goyal, Chaudhary, Wenzek, Guzm{\'a}n, Grave, Ott, Zettlemoyer, and Stoyanov}]{conneau2020unsupervised}
Alexis Conneau, Kartikay Khandelwal, Naman Goyal, Vishrav Chaudhary, Guillaume Wenzek, Francisco Guzm{\'a}n, {\'E}douard Grave, Myle Ott, Luke Zettlemoyer, and Veselin Stoyanov. 2020.
\newblock Unsupervised cross-lingual representation learning at scale.
\newblock In \emph{Proceedings of the 58th Annual Meeting of the Association for Computational Linguistics}, pages 8440--8451.

\bibitem[{Costa-juss{\`a} et~al.(2022)Costa-juss{\`a}, Cross, {\c{C}}elebi, Elbayad, Heafield, Heffernan, Kalbassi, Lam, Licht, Maillard et~al.}]{costa2022no}
Marta~R Costa-juss{\`a}, James Cross, Onur {\c{C}}elebi, Maha Elbayad, Kenneth Heafield, Kevin Heffernan, Elahe Kalbassi, Janice Lam, Daniel Licht, Jean Maillard, et~al. 2022.
\newblock No language left behind: Scaling human-centered machine translation.
\newblock \emph{arXiv preprint arXiv:2207.04672}.

\bibitem[{Devlin et~al.(2019)Devlin, Chang, Lee, and Toutanova}]{devlin2019bert}
Jacob Devlin, Ming-Wei Chang, Kenton Lee, and Kristina Toutanova. 2019.
\newblock Bert: Pre-training of deep bidirectional transformers for language understanding.
\newblock In \emph{Proceedings of the 2019 Conference of the North American Chapter of the Association for Computational Linguistics: Human Language Technologies, Volume 1 (Long and Short Papers)}, pages 4171--4186.

\bibitem[{Dou et~al.(2024)Dou, Zhou, Liu, Gao, Zhao, Shen, Zhou, Xi, Wang, Fan, Pu, Zhu, Zheng, Gui, Zhang, and Huang}]{dou2024loramoe}
Shihan Dou, Enyu Zhou, Yan Liu, Songyang Gao, Jun Zhao, Wei Shen, Yuhao Zhou, Zhiheng Xi, Xiao Wang, Xiaoran Fan, Shiliang Pu, Jiang Zhu, Rui Zheng, Tao Gui, Qi~Zhang, and Xuanjing Huang. 2024.
\newblock \href {https://arxiv.org/abs/2312.09979} {Loramoe: Alleviate world knowledge forgetting in large language models via moe-style plugin}.
\newblock \emph{Preprint}, arXiv:2312.09979.

\bibitem[{Es et~al.(2023)Es, James, Espinosa-Anke, and Schockaert}]{es2023ragas}
Shahul Es, Jithin James, Luis Espinosa-Anke, and Steven Schockaert. 2023.
\newblock Ragas: Automated evaluation of retrieval augmented generation.
\newblock \emph{arXiv preprint arXiv:2309.15217}.

\bibitem[{Fan et~al.(2021{\natexlab{a}})Fan, Bhosale, Schwenk, Ma, El-Kishky, Goyal, Baines, Celebi, Wenzek, Chaudhary et~al.}]{fan2021beyond}
Angela Fan, Shruti Bhosale, Holger Schwenk, Zhiyi Ma, Ahmed El-Kishky, Siddharth Goyal, Mandeep Baines, Onur Celebi, Guillaume Wenzek, Vishrav Chaudhary, et~al. 2021{\natexlab{a}}.
\newblock Beyond english-centric multilingual machine translation.
\newblock \emph{Journal of Machine Learning Research}, 22(107):1--48.

\bibitem[{Fan et~al.(2021{\natexlab{b}})Fan, Liang, Muzio, Awadalla, Li, Zhou, and Duan}]{fan2021discovering}
Yimin Fan, Yaobo Liang, Alexandre Muzio, Hany~Hassan Awadalla, Houqiang Li, Ming Zhou, and Nan Duan. 2021{\natexlab{b}}.
\newblock Discovering representation sprachbund for multilingual pre-training.
\newblock In \emph{Findings of the Association for Computational Linguistics: EMNLP 2021}, pages 881--894.

\bibitem[{Gupta and Srikumar(2021)}]{gupta2021x}
Ashim Gupta and Vivek Srikumar. 2021.
\newblock X-fact: A new benchmark dataset for multilingual fact checking.
\newblock \emph{arXiv preprint arXiv:2106.09248}.

\bibitem[{Huang et~al.(2019)Huang, Liang, Duan, Gong, Shou, Jiang, and Zhou}]{huang-etal-2019-unicoder}
Haoyang Huang, Yaobo Liang, Nan Duan, Ming Gong, Linjun Shou, Daxin Jiang, and Ming Zhou. 2019.
\newblock \href {https://doi.org/10.18653/v1/D19-1252} {{U}nicoder: A universal language encoder by pre-training with multiple cross-lingual tasks}.
\newblock In \emph{Proceedings of the 2019 Conference on Empirical Methods in Natural Language Processing and the 9th International Joint Conference on Natural Language Processing (EMNLP-IJCNLP)}, pages 2485--2494, Hong Kong, China. Association for Computational Linguistics.

\bibitem[{Jiang et~al.(2023)Jiang, Sablayrolles, Mensch, Bamford, Chaplot, Casas, Bressand, Lengyel, Lample, Saulnier et~al.}]{jiang2023mistral}
Albert~Q Jiang, Alexandre Sablayrolles, Arthur Mensch, Chris Bamford, Devendra~Singh Chaplot, Diego de~las Casas, Florian Bressand, Gianna Lengyel, Guillaume Lample, Lucile Saulnier, et~al. 2023.
\newblock Mistral 7b.
\newblock \emph{arXiv preprint arXiv:2310.06825}.

\bibitem[{Ju et~al.(2024)Ju, Sun, Du, Yuan, Ren, and Liu}]{ju2024large}
Tianjie Ju, Weiwei Sun, Wei Du, Xinwei Yuan, Zhaochun Ren, and Gongshen Liu. 2024.
\newblock How large language models encode context knowledge? a layer-wise probing study.
\newblock \emph{arXiv preprint arXiv:2402.16061}.

\bibitem[{Kulshreshtha et~al.(2020)Kulshreshtha, Garcia, and Chang}]{kulshreshtha2020cross}
Saurabh Kulshreshtha, Jose Luis~Redondo Garcia, and Ching~Yun Chang. 2020.
\newblock Cross-lingual alignment methods for multilingual bert: A comparative study.
\newblock In \emph{Findings of the Association for Computational Linguistics: EMNLP 2020}, pages 933--942.

\bibitem[{Lai et~al.(2023)Lai, Nguyen, Ngo, Nguyen, Dernoncourt, Rossi, and Nguyen}]{lai2023okapi}
Viet Lai, Chien Nguyen, Nghia Ngo, Thuat Nguyen, Franck Dernoncourt, Ryan Rossi, and Thien Nguyen. 2023.
\newblock Okapi: Instruction-tuned large language models in multiple languages with reinforcement learning from human feedback.
\newblock In \emph{Proceedings of the 2023 Conference on Empirical Methods in Natural Language Processing: System Demonstrations}, pages 318--327.

\bibitem[{Lample and Conneau(2019)}]{lample2019cross}
Guillaume Lample and Alexis Conneau. 2019.
\newblock Cross-lingual language model pretraining.
\newblock \emph{arXiv preprint arXiv:1901.07291}.

\bibitem[{Lewis et~al.(2020)Lewis, Oguz, Rinott, Riedel, and Schwenk}]{lewis2020mlqa}
Patrick Lewis, Barlas Oguz, Ruty Rinott, Sebastian Riedel, and Holger Schwenk. 2020.
\newblock Mlqa: Evaluating cross-lingual extractive question answering.
\newblock In \emph{Proceedings of the 58th Annual Meeting of the Association for Computational Linguistics}, pages 7315--7330.

\bibitem[{Li et~al.(2024{\natexlab{a}})Li, Alkhouli, Bonadiman, Pappas, and Mansour}]{li2024eliciting}
Bryan Li, Tamer Alkhouli, Daniele Bonadiman, Nikolaos Pappas, and Saab Mansour. 2024{\natexlab{a}}.
\newblock Eliciting better multilingual structured reasoning from llms through code.
\newblock \emph{arXiv preprint arXiv:2403.02567}.

\bibitem[{Li et~al.(2024{\natexlab{b}})Li, Haider, and Callison-Burch}]{li-etal-2024-land}
Bryan Li, Samar Haider, and Chris Callison-Burch. 2024{\natexlab{b}}.
\newblock \href {https://doi.org/10.18653/v1/2024.naacl-long.213} {This land is {Your, My} land: Evaluating geopolitical bias in language models through territorial disputes}.
\newblock In \emph{Proceedings of the 2024 Conference of the North American Chapter of the Association for Computational Linguistics: Human Language Technologies (Volume 1: Long Papers)}, pages 3855--3871, Mexico City, Mexico. Association for Computational Linguistics.

\bibitem[{Lin et~al.(2024)Lin, Wang, Liu, and Chen}]{lin2024crossin}
Geyu Lin, Bin Wang, Zhengyuan Liu, and Nancy~F Chen. 2024.
\newblock Crossin: An efficient instruction tuning approach for cross-lingual knowledge alignment.
\newblock \emph{arXiv preprint arXiv:2404.11932}.

\bibitem[{Lin et~al.(2023)Lin, Hu, Zhang, Martins, and Sch{\"u}tze}]{lin2023mplm}
Peiqin Lin, Chengzhi Hu, Zheyu Zhang, Andr{\'e}~FT Martins, and Hinrich Sch{\"u}tze. 2023.
\newblock mplm-sim: Unveiling better cross-lingual similarity and transfer in multilingual pretrained language models.
\newblock \emph{arXiv preprint arXiv:2305.13684}.

\bibitem[{Lin et~al.(2022)Lin, Hilton, and Evans}]{lin2022truthfulqa}
Stephanie Lin, Jacob Hilton, and Owain Evans. 2022.
\newblock Truthfulqa: Measuring how models mimic human falsehoods.
\newblock In \emph{Proceedings of the 60th Annual Meeting of the Association for Computational Linguistics (Volume 1: Long Papers)}, pages 3214--3252.

\bibitem[{Maynez et~al.(2020)Maynez, Narayan, Bohnet, and McDonald}]{maynez-etal-2020-faithfulness}
Joshua Maynez, Shashi Narayan, Bernd Bohnet, and Ryan McDonald. 2020.
\newblock \href {https://doi.org/10.18653/v1/2020.acl-main.173} {On faithfulness and factuality in abstractive summarization}.
\newblock In \emph{Proceedings of the 58th Annual Meeting of the Association for Computational Linguistics}, pages 1906--1919, Online. Association for Computational Linguistics.

\bibitem[{Min et~al.(2023)Min, Krishna, Lyu, Lewis, Yih, Koh, Iyyer, Zettlemoyer, and Hajishirzi}]{min2023factscore}
Sewon Min, Kalpesh Krishna, Xinxi Lyu, Mike Lewis, Wen-tau Yih, Pang Koh, Mohit Iyyer, Luke Zettlemoyer, and Hannaneh Hajishirzi. 2023.
\newblock Factscore: Fine-grained atomic evaluation of factual precision in long form text generation.
\newblock In \emph{Proceedings of the 2023 Conference on Empirical Methods in Natural Language Processing}, pages 12076--12100.

\bibitem[{Muennighoff et~al.(2023)Muennighoff, Wang, Sutawika, Roberts, Biderman, Le~Scao, Bari, Shen, Yong, Schoelkopf et~al.}]{muennighoff2023crosslingual}
Niklas Muennighoff, Thomas Wang, Lintang Sutawika, Adam Roberts, Stella Biderman, Teven Le~Scao, M~Saiful Bari, Sheng Shen, Zheng~Xin Yong, Hailey Schoelkopf, et~al. 2023.
\newblock Crosslingual generalization through multitask finetuning.
\newblock In \emph{Proceedings of the 61st Annual Meeting of the Association for Computational Linguistics (Volume 1: Long Papers)}, pages 15991--16111.

\bibitem[{Qi et~al.(2023)Qi, Fern{\'a}ndez, and Bisazza}]{qi2023cross}
Jirui Qi, Raquel Fern{\'a}ndez, and Arianna Bisazza. 2023.
\newblock Cross-lingual consistency of factual knowledge in multilingual language models.
\newblock \emph{arXiv preprint arXiv:2310.10378}.

\bibitem[{Qin et~al.(2024)Qin, Chen, Zhou, Chen, Li, Liao, Li, Che, and Yu}]{qin2024multilingual}
Libo Qin, Qiguang Chen, Yuhang Zhou, Zhi Chen, Yinghui Li, Lizi Liao, Min Li, Wanxiang Che, and Philip~S Yu. 2024.
\newblock Multilingual large language model: A survey of resources, taxonomy and frontiers.
\newblock \emph{arXiv preprint arXiv:2404.04925}.

\bibitem[{Qiu et~al.(2023)Qiu, Ziser, Korhonen, Ponti, and Cohen}]{qiu2023detecting}
Yifu Qiu, Yftah Ziser, Anna Korhonen, Edoardo~M Ponti, and Shay~B Cohen. 2023.
\newblock Detecting and mitigating hallucinations in multilingual summarisation.
\newblock \emph{arXiv preprint arXiv:2305.13632}.

\bibitem[{Sanders(1987)}]{sanders1987pareto}
Robert Sanders. 1987.
\newblock The pareto principle: its use and abuse.
\newblock \emph{Journal of Services Marketing}, 1(2):37--40.

\bibitem[{Schwenk et~al.(2021)Schwenk, Chaudhary, Sun, Gong, and Guzm{\'a}n}]{schwenk2021wikimatrix}
Holger Schwenk, Vishrav Chaudhary, Shuo Sun, Hongyu Gong, and Francisco Guzm{\'a}n. 2021.
\newblock Wikimatrix: Mining 135m parallel sentences in 1620 language pairs from wikipedia.
\newblock In \emph{Proceedings of the 16th Conference of the European Chapter of the Association for Computational Linguistics: Main Volume}, pages 1351--1361.

\bibitem[{Shafayat et~al.(2024)Shafayat, Kim, Oh, and Oh}]{shafayat2024multi}
Sheikh Shafayat, Eunsu Kim, Juhyun Oh, and Alice Oh. 2024.
\newblock Multi-fact: Assessing multilingual llms' multi-regional knowledge using factscore.
\newblock \emph{arXiv preprint arXiv:2402.18045}.

\bibitem[{Shi et~al.(2022)Shi, Suzgun, Freitag, Wang, Srivats, Vosoughi, Chung, Tay, Ruder, Zhou et~al.}]{shi2022language}
Freda Shi, Mirac Suzgun, Markus Freitag, Xuezhi Wang, Suraj Srivats, Soroush Vosoughi, Hyung~Won Chung, Yi~Tay, Sebastian Ruder, Denny Zhou, et~al. 2022.
\newblock Language models are multilingual chain-of-thought reasoners.
\newblock \emph{arXiv preprint arXiv:2210.03057}.

\bibitem[{Shi et~al.(2023)Shi, Han, Lewis, Tsvetkov, Zettlemoyer, and Yih}]{shi2023trusting}
Weijia Shi, Xiaochuang Han, Mike Lewis, Yulia Tsvetkov, Luke Zettlemoyer, and Scott Wen-tau Yih. 2023.
\newblock Trusting your evidence: Hallucinate less with context-aware decoding.
\newblock \emph{arXiv preprint arXiv:2305.14739}.

\bibitem[{Sun et~al.(2024)Sun, Huang, Wang, Wu, Zhang, Gao, Huang, Lyu, Zhang, Li et~al.}]{sun2024trustllm}
Lichao Sun, Yue Huang, Haoran Wang, Siyuan Wu, Qihui Zhang, Chujie Gao, Yixin Huang, Wenhan Lyu, Yixuan Zhang, Xiner Li, et~al. 2024.
\newblock Trustllm: Trustworthiness in large language models.
\newblock \emph{arXiv preprint arXiv:2401.05561}.

\bibitem[{Tan et~al.(2019)Tan, Chen, He, Xia, Qin, and Liu}]{tan2019multilingual}
Xu~Tan, Jiale Chen, Di~He, Yingce Xia, Tao Qin, and Tie-Yan Liu. 2019.
\newblock Multilingual neural machine translation with language clustering.
\newblock In \emph{Proceedings of the 2019 Conference on Empirical Methods in Natural Language Processing and the 9th International Joint Conference on Natural Language Processing (EMNLP-IJCNLP)}, pages 963--973.

\bibitem[{Taori et~al.(2023)Taori, Gulrajani, Zhang, Dubois, Li, Guestrin, Liang, and Hashimoto}]{alpaca}
Rohan Taori, Ishaan Gulrajani, Tianyi Zhang, Yann Dubois, Xuechen Li, Carlos Guestrin, Percy Liang, and Tatsunori~B. Hashimoto. 2023.
\newblock Stanford alpaca: An instruction-following llama model.
\newblock \url{https://github.com/tatsu-lab/stanford_alpaca}.

\bibitem[{Team et~al.(2023)Team, Anil, Borgeaud, Wu, Alayrac, Yu, Soricut, Schalkwyk, Dai, Hauth et~al.}]{team2023gemini}
Gemini Team, Rohan Anil, Sebastian Borgeaud, Yonghui Wu, Jean-Baptiste Alayrac, Jiahui Yu, Radu Soricut, Johan Schalkwyk, Andrew~M Dai, Anja Hauth, et~al. 2023.
\newblock Gemini: a family of highly capable multimodal models.
\newblock \emph{arXiv preprint arXiv:2312.11805}.

\bibitem[{Team et~al.(2024)Team, Mesnard, Hardin, Dadashi, Bhupatiraju, Pathak, Sifre, Rivi{\`e}re, Kale, Love et~al.}]{team2024gemma}
Gemma Team, Thomas Mesnard, Cassidy Hardin, Robert Dadashi, Surya Bhupatiraju, Shreya Pathak, Laurent Sifre, Morgane Rivi{\`e}re, Mihir~Sanjay Kale, Juliette Love, et~al. 2024.
\newblock Gemma: Open models based on gemini research and technology.
\newblock \emph{arXiv preprint arXiv:2403.08295}.

\bibitem[{Tian et~al.(2023)Tian, Mitchell, Yao, Manning, and Finn}]{tian2023fine}
Katherine Tian, Eric Mitchell, Huaxiu Yao, Christopher~D Manning, and Chelsea Finn. 2023.
\newblock Fine-tuning language models for factuality.
\newblock \emph{arXiv preprint arXiv:2311.08401}.

\bibitem[{Tiedemann(2020)}]{tiedemann2020tatoeba}
J{\"o}rg Tiedemann. 2020.
\newblock The tatoeba translation challenge--realistic data sets for low resource and multilingual mt.
\newblock In \emph{Proceedings of the Fifth Conference on Machine Translation}, pages 1174--1182.

\bibitem[{Tonmoy et~al.(2024)Tonmoy, Zaman, Jain, Rani, Rawte, Chadha, and Das}]{tonmoy2024comprehensive}
SM~Tonmoy, SM~Zaman, Vinija Jain, Anku Rani, Vipula Rawte, Aman Chadha, and Amitava Das. 2024.
\newblock A comprehensive survey of hallucination mitigation techniques in large language models.
\newblock \emph{arXiv preprint arXiv:2401.01313}.

\bibitem[{Touvron et~al.(2023)Touvron, Martin, Stone, Albert, Almahairi, Babaei, Bashlykov, Batra, Bhargava, Bhosale et~al.}]{touvron2023llama}
Hugo Touvron, Louis Martin, Kevin Stone, Peter Albert, Amjad Almahairi, Yasmine Babaei, Nikolay Bashlykov, Soumya Batra, Prajjwal Bhargava, Shruti Bhosale, et~al. 2023.
\newblock Llama 2: Open foundation and fine-tuned chat models.
\newblock \emph{arXiv preprint arXiv:2307.09288}.

\bibitem[{Vu et~al.(2023)Vu, Iyyer, Wang, Constant, Wei, Wei, Tar, Sung, Zhou, Le et~al.}]{vu2023freshllms}
Tu~Vu, Mohit Iyyer, Xuezhi Wang, Noah Constant, Jerry Wei, Jason Wei, Chris Tar, Yun-Hsuan Sung, Denny Zhou, Quoc Le, et~al. 2023.
\newblock Freshllms: Refreshing large language models with search engine augmentation.
\newblock \emph{arXiv preprint arXiv:2310.03214}.

\bibitem[{Wang et~al.(2023{\natexlab{a}})Wang, Liu, Huang, Jiao, Ding, Aw, and Chen}]{wang2023seaeval}
Bin Wang, Zhengyuan Liu, Xin Huang, Fangkai Jiao, Yang Ding, Ai~Ti Aw, and Nancy~F Chen. 2023{\natexlab{a}}.
\newblock Seaeval for multilingual foundation models: From cross-lingual alignment to cultural reasoning.
\newblock \emph{arXiv preprint arXiv:2309.04766}.

\bibitem[{Wang et~al.(2023{\natexlab{b}})Wang, Liu, Yue, Tang, Zhang, Jiayang, Yao, Gao, Hu, Qi et~al.}]{wang2023survey}
Cunxiang Wang, Xiaoze Liu, Yuanhao Yue, Xiangru Tang, Tianhang Zhang, Cheng Jiayang, Yunzhi Yao, Wenyang Gao, Xuming Hu, Zehan Qi, et~al. 2023{\natexlab{b}}.
\newblock Survey on factuality in large language models: Knowledge, retrieval and domain-specificity.
\newblock \emph{arXiv preprint arXiv:2310.07521}.

\bibitem[{Xue et~al.(2021)Xue, Constant, Roberts, Kale, Al-Rfou, Siddhant, Barua, and Raffel}]{xue2021mt5}
Linting Xue, Noah Constant, Adam Roberts, Mihir Kale, Rami Al-Rfou, Aditya Siddhant, Aditya Barua, and Colin Raffel. 2021.
\newblock mt5: A massively multilingual pre-trained text-to-text transformer.
\newblock In \emph{Proceedings of the 2021 Conference of the North American Chapter of the Association for Computational Linguistics: Human Language Technologies}, pages 483--498.

\bibitem[{Yang et~al.(2022)Yang, Yin, Ma, Zhang, Li, and Wei}]{yang2022hlt}
Jian Yang, Yuwei Yin, Shuming Ma, Dongdong Zhang, Zhoujun Li, and Furu Wei. 2022.
\newblock Hlt-mt: High-resource language-specific training for multilingual neural machine translation.
\newblock \emph{arXiv preprint arXiv:2207.04906}.

\bibitem[{Zhang et~al.(2024)Zhang, Yu, and Feng}]{zhang2024truthx}
Shaolei Zhang, Tian Yu, and Yang Feng. 2024.
\newblock Truthx: Alleviating hallucinations by editing large language models in truthful space.
\newblock \emph{arXiv preprint arXiv:2402.17811}.

\bibitem[{Zhang et~al.(2023{\natexlab{a}})Zhang, Dong, Li, Zhang, Sun, Wang, Li, Hu, Zhang, Wu et~al.}]{zhang2023instruction}
Shengyu Zhang, Linfeng Dong, Xiaoya Li, Sen Zhang, Xiaofei Sun, Shuhe Wang, Jiwei Li, Runyi Hu, Tianwei Zhang, Fei Wu, et~al. 2023{\natexlab{a}}.
\newblock Instruction tuning for large language models: A survey.
\newblock \emph{arXiv preprint arXiv:2308.10792}.

\bibitem[{Zhang et~al.(2021)Zhang, Ma, Shi, and Lin}]{mrtydi}
Xinyu Zhang, Xueguang Ma, Peng Shi, and Jimmy Lin. 2021.
\newblock {Mr. TyDi}: A multi-lingual benchmark for dense retrieval.
\newblock \emph{arXiv:2108.08787}.

\bibitem[{Zhang et~al.(2023{\natexlab{b}})Zhang, Li, Cui, Cai, Liu, Fu, Huang, Zhao, Zhang, Chen, Wang, Luu, Bi, Shi, and Shi}]{zhang2023sirens}
Yue Zhang, Yafu Li, Leyang Cui, Deng Cai, Lemao Liu, Tingchen Fu, Xinting Huang, Enbo Zhao, Yu~Zhang, Yulong Chen, Longyue Wang, Anh~Tuan Luu, Wei Bi, Freda Shi, and Shuming Shi. 2023{\natexlab{b}}.
\newblock \href {https://arxiv.org/abs/2309.01219} {Siren's song in the ai ocean: A survey on hallucination in large language models}.
\newblock \emph{Preprint}, arXiv:2309.01219.

\bibitem[{Zhu et~al.(2023)Zhu, Lv, Dong, Yuan, Xu, Huang, Kong, Chen, and Li}]{zhu2023extrapolating}
Wenhao Zhu, Yunzhe Lv, Qingxiu Dong, Fei Yuan, Jingjing Xu, Shujian Huang, Lingpeng Kong, Jiajun Chen, and Lei Li. 2023.
\newblock Extrapolating large language models to non-english by aligning languages.
\newblock \emph{arXiv preprint arXiv:2308.04948}.

\bibitem[{Ziemski et~al.(2016)Ziemski, Junczys-Dowmunt, and Pouliquen}]{ziemski2016united}
Micha{\l} Ziemski, Marcin Junczys-Dowmunt, and Bruno Pouliquen. 2016.
\newblock The united nations parallel corpus v1. 0.
\newblock In \emph{Proceedings of the Tenth International Conference on Language Resources and Evaluation (LREC'16)}, pages 3530--3534.

\end{thebibliography}

\appendix



\section{Details of Benchmark Construction}

\subsection{Prompt for Translation}
\label{app:mtfqa_trans}

\begin{figure}[htbp]
    \centering
    \includegraphics[width= 0.5\textwidth]{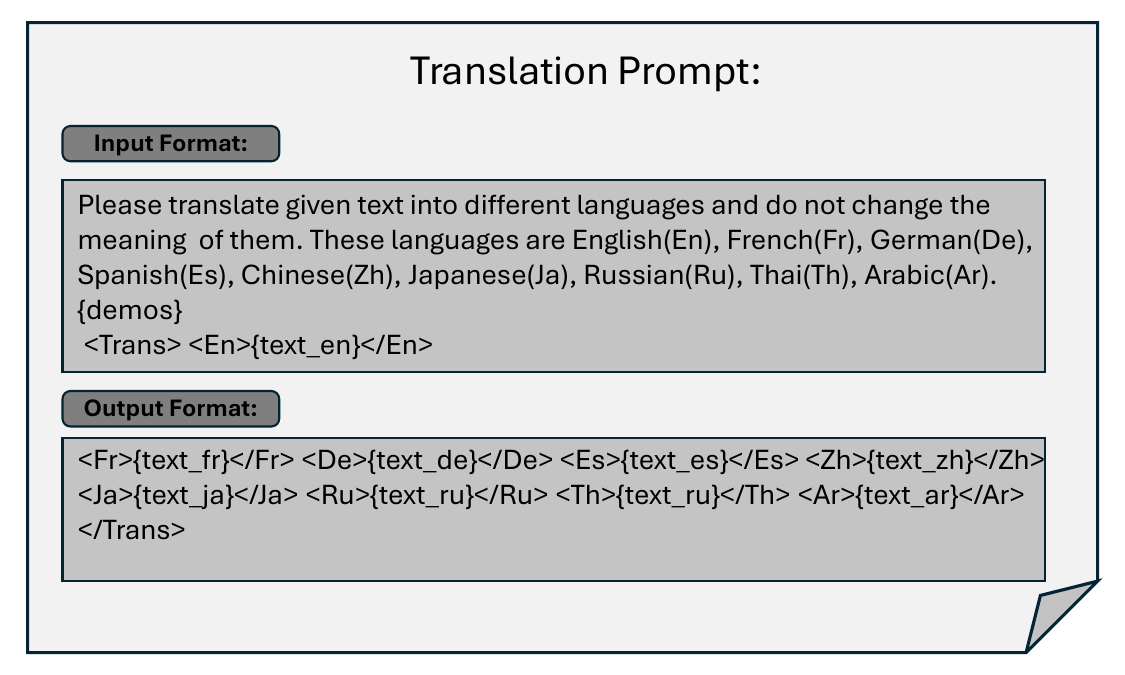}
    \caption{Prompt to translate questions and answers.}
    \label{fig:prompt_benchmark}
\end{figure}

We use the prompt format in Figure~\ref{fig:prompt_benchmark} to translate questions and answers in TruthfulQA into other languages.


\subsection{Metrics for MTruthfulQA}
\label{app:metrics}

Formally, we define the question in language $l$ as $\mathcal{Q}_l$ and the response of LLM $\mathcal{L}$ as $\mathcal{R_L}(\mathcal{Q}_l)$. For multi-choice QA, we define $\mathcal{AB}_l$ as the best answer, $\mathcal{AT}_l$ as the set of correct answers and $\mathcal{AF}_l$ as the set of wrong answers in corresponding language $l$. The three likelihood scores $MC1$, $MC2$ and $MC3$ are calculated as follows:
$$
MC1_l = \begin{cases}1, & if\ p(\mathcal{AB}_l)>\max_i p(\mathcal{AF}_{l,i}) \\ 0, & else\end{cases}
$$

$$
MC2_l = \dfrac{\sum_i \exp(p(\mathcal{AT}_{l,i}))}{\sum_i\exp(p(\mathcal{AB}_{l,i})) + \sum_i \exp(p(\mathcal{AF}_{l,i}))}
$$

$$
MC3_l = \dfrac{\sum_i [p(\mathcal{AT}_{l,i})>p(\mathcal{AF}_{l,i})]}{|\mathcal{AB}_l|}
$$

where $l$ represents the language evaluated on, $p(x)$ represents the logits of $x$ and the symbol with subscript $i$ denotes the $i$th item in the corresponding set. The final $MC$ scores are the average scores of all test samples.

For open-ended generation:
\begin{itemize}
    \item \textbf{True~(\%)}: the percentage of responses that are classified as truthful.
    \item \textbf{Info~(\%)}: the percentage of responses that are classified as informative.
    \item \textbf{True*Info~(\%)}: the percentage of responses that are classified as both truthful and informative.
\end{itemize}

\section{Details of the Fact-aware Multilingual Alignment}
\label{app:translate_detail}

\subsection{Hyperparameters}

\begin{table}[H]
\begin{tabular}{ll}
\toprule
\textbf{Hyperparameter}     & \textbf{Value} \\ \hline
learning\_rate               & 3e-6           \\ \hline
batch\_size                  & 4              \\ \hline
gradient\_accumulation\_steps & 2              \\ \hline
epochs                      & 4              \\ \hline
model\_max\_length            & 2048           \\ \hline
lr\_scheduler\_type           & cosine         \\ \hline
fp16                        & True           \\ \hline
optimizer                   & AdamW          \\ \bottomrule
\end{tabular}
\caption{Fine-tuning hyperparameters.}
\label{tab:hyperparameters}
\end{table}

\subsection{Statistics of Training Data}
\label{app:exampledata}

Table~\ref{tab:all_training_sta} shows the statistics on the two types of translation data we collect. Table~\ref{tab:detailed_training_data} presents the allocation of training data in our main experiment.

\begin{table}[htb]
\centering
\resizebox{0.48 \textwidth}{!}{
\begin{tabular}{cccccccccc}
\toprule
Data Type              & De & Fr & Es & Ru & Zh & Ja & Th & Ar \\ \hline
Factuality Translation & 4517              & 4235              & 5253              & 5223              & 5137              & 4236              & 4239              & 5335              \\ \hline
Common Translation     & 997               & 997               & 997               & 997               & 997               & 997               & 997               & 997               \\ 
\bottomrule
\end{tabular}
}
\caption{Statistics of collected training data in our experiments.}
\label{tab:all_training_sta}
\end{table}

\begin{table}[h]
\centering
\resizebox{0.48 \textwidth}{!}{
\begin{tabular}{ccc}
\toprule
\textbf{Data Type}                                      & \textbf{Target Language} & \textbf{Number of Items} \\ \hline
\multirow{3}{*}{Factuality Translation}                 & De           & 4517                     \\
                                                        & Zh           & 5137                     \\
                                                        & Ar           & 5335                     \\ \hline
\multicolumn{1}{l}{\multirow{3}{*}{Common Translation}} & De           & 997                      \\
\multicolumn{1}{l}{}                                    & Zh           & 997                      \\
\multicolumn{1}{l}{}                                    & Ar           & 997                      \\ \hline
Pretraining                                             & En                          & 1946                     \\
\bottomrule
\end{tabular}
}
\caption{Detailed information of training data used for \FMT~with optimal allocation of data types and languages.}
\label{tab:detailed_training_data}
\end{table}

\subsection{Template for Translation Instruction Tuning}
\label{app:training_format}
\begin{figure}[H]
    \centering
    \includegraphics[width= 0.5\textwidth]{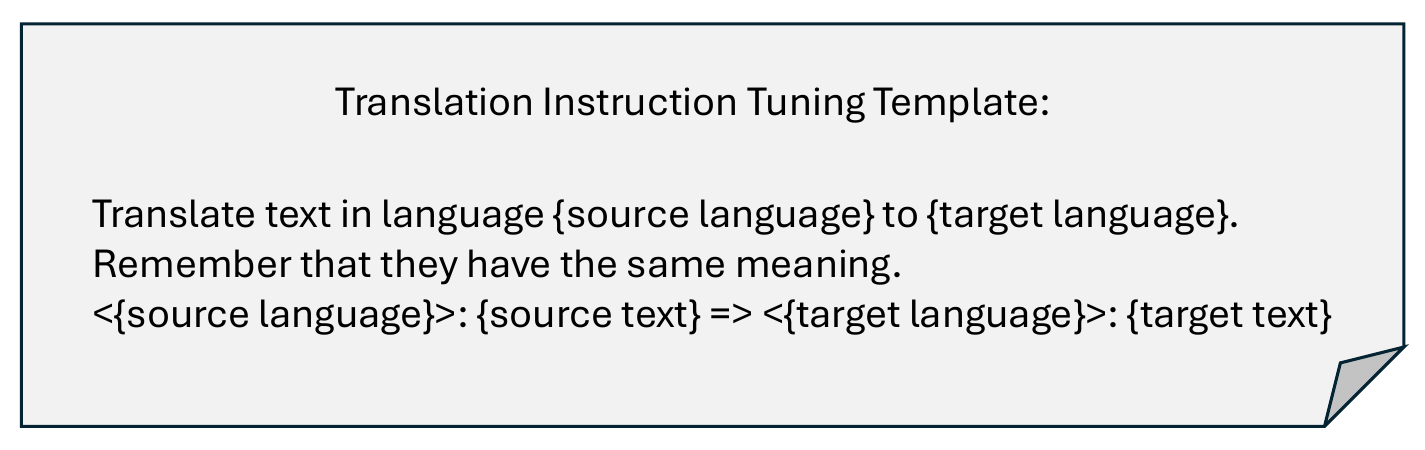}
    \caption{Format of instruction tuning data in translation task.}
    \label{fig:training_format}
\end{figure}

\section{Details About Optimal Languages Selection}
\label{app:detail_selection}

\begin{figure*}[ht]
    \centering
    \includegraphics[width = \textwidth]{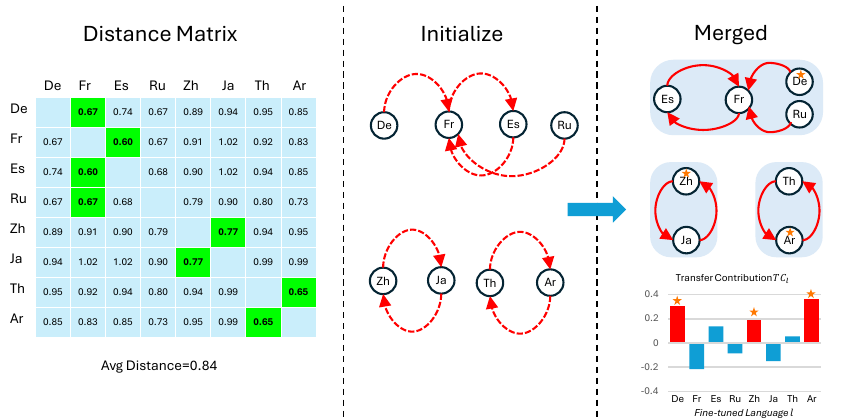}
    \caption{Process of selecting optimal language set. \textbf{Left:} distance matrix between each language pair. The numbers in green are the minimum values of that row. \textbf{Middle:} initialization and confirm the nearest language for each language. Edge from X to Y means the nearest language of X is Y. \textbf{Right:} three language sets after merging. Starred language is the core language of that set, which is determined by transfer contribution $\mathcal{TC}_l$.}
    \label{fig:select}
\end{figure*}

In the process of optimal language set selection described in Algorithm~\ref{algo:lan_select}, $M$ is set to 3 and $d$ is set to the average bias between all different language pairs (i.e., $d=0.84$). As shown in Figure~\ref{fig:select}, we first merge all languages into three groups, each containing similar languages. Then we calculate the transfer contribution $\mathcal{TC}_l$ according to the mean bias movement for each language. We finally select three core languages from these corresponding three sets~(i.e., German, Chinese and Arabic).

\section{More Results}
\label{app:result}
We report more detailed results evaluated on MTruthfulQA in Table~\ref{tab:true_score},~\ref{tab:info_score},~\ref{tab:mc2_score} and~\ref{tab:mc3_score}. These metrics include True~(\%), Info~(\%), $MC2$~(\%) and $MC3$~(\%). We also present the ablation studies results over different $m$, $d$ values in Table~\ref{tab:dif_md}.




\begin{table*}[]
\centering
\begin{tabular}{l|cccccccccc}
\toprule
\multirow{2}{*}{\textbf{Models}}                            & \multicolumn{10}{c}{\textbf{True~(\%)}}                                                                                                      \\ \cline{2-11} 
                                                            & \textbf{En} & \textbf{De} & \textbf{Fr} & \textbf{Es} & \textbf{Ru} & \textbf{Zh} & \textbf{Ja} & \textbf{Th} & \textbf{Ar} & \textbf{Avg.} \\ \hline
\textit{GPT-4}                                              & 72.2        & 72.7        & 73.7        & 69.3        & 70.4        & 70.4        & 68.9        & 57.0        & 71.5        & 69.6          \\
\textit{Bloomz-7B1-mt}                                      & 92.5        & 55.0        & 92.3        & 99.5        & 84.8        & 76.1        & 82.1        & 68.4        & 91.3        & 82.5          \\
\textit{LLaMA-3-8B-Instruct}                                & 71.2        & 85.7        & 70.5        & 77.4        & 47.7        & 74.3        & 91.4        & 76.5        & 68.4        & 76.0          \\
\textit{Mistral-7B-Instruct-v0.2}                           & 80.0        & 78.6        & 76.5        & 79.2        & 75.4        & 69.8        & 80.2        & 51.9        & 69.8        & 74.1          \\ 
\hline
\textbf{With \FMT}                           &             &             &             &             &             &             &             &             &             &               \\ \hline
\textit{LLaMA-3-8B}                                         & 58.9        & 78.6        & 76.5        & 79.2        & 75.4        & 69.8        & 80.2        & 51.9        & 74.9        & 74.1          \\
\textit{LLaMA-3-8B} + \textbf{\FMT}  & 60.5        & 62.5        & 56.3        & 65.0        & 49.4        & 47.9        & 57.2        & 37.9        & 52.3        & 54.3          \\ \hline
\textit{Mistral-7B-v0.1}                                    & 55.2        & 59.0        & 60.8        & 68.3        & 61.9        & 49.7        & 82.1        & 67.1        & 81.5        & 65.1          \\
\textit{Mistral-7B-v0.1} + \textbf{\FMT}                    & 46.4        & 59.6        & 66.7        & 70.6        & 63.0        & 52.6        & 84.3        & 64.7        & 85.9        & 66.0          \\ \hline
\textit{Gemma-7B}                                        & 76.5        & 85.6        & 85.2        & 87.3        & 80.2        & 71.2        & 69.9        & 76.9        & 77.7        & 78.9          \\
\textit{Gemma-7B} + \textbf{\FMT} & 78.5        & 84.7        & 86.0        & 86.2        & 73.6        & 64.7        & 60.7        & 68.7        & 79.2        & 75.8         \\
\bottomrule
\end{tabular}
\caption{True~(\%) score of models evaluated on MTruthfulQA. In our experiment, we find that a single True~(\%) score does not reflect the actual ability to give truthful answers since models are always refusing to give a clear answer, which also improve their True~(\%) score.}
\label{tab:true_score}
\end{table*}

\begin{table*}[]
\centering
\begin{tabular}{l|cccccccccc}
\toprule
\multirow{2}{*}{\textbf{Models}}                            & \multicolumn{10}{c}{\textbf{Info~(\%)}}                                                                                                      \\ \cline{2-11} 
                                                            & \textbf{En} & \textbf{De} & \textbf{Fr} & \textbf{Es} & \textbf{Ru} & \textbf{Zh} & \textbf{Ja} & \textbf{Th} & \textbf{Ar} & \textbf{Avg.} \\ \hline
\textit{GPT-4}                                              & 97.6        & 98.4        & 98.8        & 98.7        & 98.4        & 98.5        & 99.0        & 99.1        & 97.1        & 98.4          \\
\textit{Bloomz-7B1-mt}                                      & 26.2        & 61.2        & 28.4        & 13.6        & 30.2        & 49.8        & 32.9        & 36.2        & 22.8        & 33.5          \\
\textit{LLaMA-3-8B-Instruct}                                & 84.9        & 45.0        & 72.6        & 51.7        & 69.4        & 60.0        & 27.2        & 40.3        & 62.5        & 57.1          \\
\textit{Mistral-7B-Instruct-v0.2}                           & 86.5        & 73.9        & 81.0        & 70.6        & 73.4        & 83.0        & 47.0        & 68.9        & 66.1        & 72.3          \\ 
\hline
\textbf{With \FMT}                           &             &             &             &             &             &             &             &             &             &               \\ \hline
\textit{LLaMA-3-8B}                                         & 77.6        & 67.9        & 69.5        & 60.7        & 84.1        & 78.2        & 50.7        & 70.9        & 83.0        & 71.4          \\
\textit{LLaMA-3-8B} + \textbf{\FMT}  & 84.0        & 72.2        & 78.2        & 72.1        & 92.5        & 88.1        & 81.4        & 85.1        & 88.5        & 82.5          \\ \hline
\textit{Mistral-7B-v0.1}                                    & 80.8        & 76.0        & 67.6        & 60.7        & 71.4        & 84.7        & 36.6        & 44.1        & 39.2        & 62.3          \\
\textit{Mistral-7B-v0.1}  + \textbf{\FMT}                   & 93.1        & 80.7        & 64.3        & 58.2        & 69.1        & 86.1        & 35.1        & 45.3        & 42.6        & 63.8          \\ \hline
\textit{Gemma-7B}                                        & 49.2        & 33.9        & 32.2        & 29.6        & 41.0        & 56.5        & 63.8        & 38.1        & 41.9        & 42.9          \\
\textit{Gemma-7B} + \textbf{\FMT} & 59.2        & 61.0        & 35.9        & 37.5        & 56.2        & 76.6        & 78.9        & 49.9        & 45.9        & 55.7     \\
\bottomrule
\end{tabular}
\caption{Info~(\%) score of models evaluated on MTruthfulQA. Higher Info~(\%) score usually means models provide more useful information.}
\label{tab:info_score}
\end{table*}

\begin{table*}[]
\centering
\begin{tabular}{l|cccccccccc}
\toprule
\multirow{2}{*}{\textbf{Models}}  & \multicolumn{10}{c}{\textbf{MC2~(\%)}}                                                                                                       \\ \cline{2-11} 
                                  & \textbf{En} & \textbf{De} & \textbf{Fr} & \textbf{Es} & \textbf{Ru} & \textbf{Zh} & \textbf{Ja} & \textbf{Th} & \textbf{Ar} & \textbf{Avg.} \\ \hline
\textit{Bloomz-7B1-mt}            & 46.4        & 47.5        & 45.5        & 42.4        & 50.9        & 47.4        & 51.2        & 48.0        & 48.2        & 47.5          \\
\textit{LLaMA-3-8B-Instruct}      & 59.3        & 56.8        & 60.0        & 55.5        & 56.2        & 54.7        & 54.7        & 48.0        & 54.2        & 55.5          \\
\textit{Mistral-7B-Instruct-v0.2} & 70.9        & 67.8        & 69.5        & 64.8        & 64.9        & 62.2        & 60.2        & 47.9        & 58.8        & 63.0          \\ 
\hline
\textbf{With \FMT}                &             &             &             &             &             &             &             &             &             &               \\ \hline
\textit{LLaMA-3-8B}               & 50.2        & 51.1        & 53.6        & 47.6        & 52.3        & 50.5        & 54.2        & 47.0        & 51.5        & 50.8          \\
\textit{LLaMA-3-8B} + \textbf{\FMT}        & \textbf{52.7}        & \textbf{52.1}        & 52.6        & \textbf{51.3}        & \textbf{53.0}        & \textbf{50.5}        & 53.8        & 44.3        &\textbf{51.7}        & \textbf{51.3}          \\ \hline
\textit{Mistral-7B-v0.1}                     & 45.9        & 51.4        & 51.9        & 45.1        & 52.9        & 51.6        & 53.2        & 46.7        & 50.4        & 49.9          \\
\textit{Mistral-7B-v0.1} + \textbf{\FMT}     & \textbf{46.3}        & \textbf{51.6}        & 51.3        & \textbf{45.8}        & \textbf{53.1}        & \textbf{52.4}        & 52.9        & 45.9        & \textbf{50.5}        & \textbf{50.0}          \\ \hline
\textit{Gemma-7B}              & 51.6        & 49.8        & 52.1        & 51.7        & 52.2        & 50.9        & 52.7        & 45.5        & 52.4        & 51.0          \\
\textit{Gemma-7B} + \textbf{\FMT}       & \textbf{56.8}        & \textbf{54.5}        & \textbf{54.6}        & \textbf{54.2}        & \textbf{52.9}        & \textbf{52.2}        & \textbf{54.7}        & \textbf{49.7}        & 51.8        & \textbf{53.5} \\
\bottomrule
\end{tabular}
\caption{$MC2$~(\%) score of models evaluated on MTruthfulQA.}
\label{tab:mc2_score}
\end{table*}

\begin{table*}[]
\centering
\begin{tabular}{l|cccccccccc}
\toprule
\multirow{2}{*}{\textbf{Models}}  & \multicolumn{10}{c}{\textbf{MC3~(\%)}}                                                                                                                         \\ \cline{2-11} 
                                  & \textbf{En}   & \textbf{De}   & \textbf{Fr}   & \textbf{Es}   & \textbf{Ru}   & \textbf{Zh}   & \textbf{Ja}   & \textbf{Th}   & \textbf{Ar}   & \textbf{Avg.} \\ \hline
\textit{Bloomz-7B1-mt}            & 24.1          & 24.9          & 24.3          & 22.5          & 28.0          & 26.0          & 28.7          & 25.4          & 25.8          & 25.5          \\
\textit{LLaMA-3-8B-Instruct}      & 34.4          & 32.5          & 35.6          & 32.8          & 32.8          & 33.2          & 32.1          & 26.3          & 32.0          & 32.4          \\
\textit{Mistral-7B-Instruct-v0.2} & 46.0          & 44.0          & 44.5          & 41.0          & 40.3          & 38.6          & 36.1          & 26.7          & 32.8          & 38.9          \\ 
\hline
\textbf{With \FMT}                &               &               &               &               &               &               &               &               &               &               \\ \hline
\textit{LLaMA-3-8B}               & 26.8          & 28.2          & 28.7          & 26.2          & 28.6          & 27.6          & 31.0          & 26.3          & 28.3          & 28.0          \\
\textit{LLaMA-3-8B + \textbf{\FMT}}        & \textbf{29.0} & \textbf{28.2} & 28.1          & \textbf{29.1} & 28.4          & \textbf{27.6} & \textbf{31.2} & 24.6          & \textbf{28.8} & \textbf{28.3} \\ \hline
\textit{Mistral-7B-v0.1}                   & 24.4          & 29.4          & 28.1          & 25.7          & 30.2          & 30.3          & 30.5          & 25.9          & 27.3          & 28.0          \\
\textit{Mistral-7B-v0.1} + \textbf{\FMT}   & \textbf{26.4}          & \textbf{29.5}          & 27.7          & \textbf{28.4}          & \textbf{30.5}          & \textbf{31.1}          & 30.3          & 25.1          & \textbf{28.5}          & \textbf{28.6}          \\ \hline
\textit{Gemma-7B}              & 28.0          & 28.1          & 27.8          & 28.8          & 28.1          & 29.2          & 29.1          & 25.0          & 28.4          & 28.1          \\
\textit{Gemma-7B + \textbf{\FMT}}       & \textbf{30.7} & \textbf{30.9} & \textbf{31.5} & \textbf{29.4} & \textbf{29.4} & \textbf{29.7} & \textbf{29.4} & \textbf{26.3} & \textbf{28.9} & \textbf{29.6} \\
\bottomrule
\end{tabular}
\caption{$MC3$~(\%) score of models evaluated on MTruthfulQA.}
\label{tab:mc3_score}
\end{table*}

\begin{table*}[]
\centering
\resizebox{ \textwidth}{!}{
\begin{tabular}{llllllllllll}
\toprule
\textbf{Languages}         & \textbf{Settings} & \textbf{En}   & \textbf{De}   & \textbf{Fr}   & \textbf{Es}   & \textbf{Ru}   & \textbf{Zh}   & \textbf{Ja}   & \textbf{Th}   & \textbf{Ar}   & \textbf{Avg.} \\ \hline
Ar                         & m=1, d=0          & 30.8          & 25.3          & 20.8          & 19.8          & 24.1          & 27.2          & 34.6          & 18.4          & 23.7          & 25.0          \\
Zh, Ar                     & m=2, d=0.84       & 32.9          & 25.6          & 20.3          & 17.6          & 27.1          & 35.1          & 35.6          & 19.5          & 24.5          & 26.5          \\
De, Zh, Ar                 & m=3, d=0.84       & \textbf{37.8} & \textbf{45.9} & 22.2          & \textbf{23.6} & \textbf{30.8} & \textbf{41.7} & 40.1          & 19.7          & \textbf{25.2} & \textbf{31.9} \\
De, Zh, Ja, Ar             & m=4, d=0.77       & 33.7          & 36.5          & 23.5          & 20.3          & 26.7          & 39.8          & 37.2          & 19.5          & 18.5          & 28.4          \\
De, Es, Zh, Ja, Ar         & m=5, d=0.67       & 34.2          & 43.1          & 24.5          & 21.4          & 24.4          & 40.1          & 40.0          & 17.4          & 18.5          & 29.3          \\
De, Es, Ru, Zh, Ja, Ar     & m=6, d=0.66       & 30.2          & 34.4          & 18.6          & 18.5          & 25.7          & 39.9          & \textbf{42.2} & 16.0          & 19.2          & 27.2          \\
De, Es, Ru, Zh, Ja, Th, Ar & m=7, d=0.65       & 28.6          & 26.1          & 18.1          & 19.0          & 19.3          & 41.4          & 37.1          & \textbf{20.6} & 20.1          & 25.6          \\
All languages              & m=8, d=0.65       & 29.0          & 26.6          & \textbf{25.3} & 18.5          & 19.5          & 41.6          & 39.1          & 18.5          & 19.3          & 26.3       \\  
\bottomrule
\end{tabular}
}
\caption{Ablation study regarding maximum number of languages $m$ and distance threshold $d$. Presented metrics are True*Info(\%) scores. It can be observed that once the number of languages exceeds a certain threshold, adding more languages does not enhance overall multilingual capability and may even impair the performance of other languages.}
\label{tab:dif_md}
\end{table*}

\section{Reliability of "MM-Judge"}
\label{app:mm-judge}
We deploy Mistral-7B-Instruct-v0.2~\cite{jiang2023mistral} model as the base model of the evaluator used for open-ended generation. One of the reasons is that this model exhibits the best performance in multi-choice QA task shown in Table~\ref{tab:MC}, which demonstrates that this model is relatively more truthful and strong in distinguishing truthful responses. After fine-tuning the model into two "MM-Judge" models, we also tested their accuracy in judging whether a response is truthful or informative. First, we tested the accuracy of truthful judgement using the correct and incorrect answers provided by MTruthfulQA dataset (we did not test informativeness at this step, since no exact labels of informativeness were provided). The results is listed in Table~\ref{tab:evaluator1}. We also tested the evaluation results of real responses from our main experiments. Specifically, we used all responses generated by \textbf{Llama-3-8B+\FMT}, then we asked a few trained annotators to assess whether the evaluator model's evaluations were correct. The results we obtained are presented in Table~\ref{tab:evaluator2}. The average accuracy of "MM-Judge" is 91.8\% in truthfulness and 97.5\% in informativeness, which is reasonably robust and reliable.

\begin{table*}[]
\centering
\begin{tabular}{lllllllll}
\toprule
\textbf{En}   & \textbf{Fr}   & \textbf{De}   & \textbf{Es}   & \textbf{Zh}   & \textbf{Ja}   & \textbf{Ru} & \textbf{Th} & \textbf{Ar} \\ \hline
99.6 & 99.1 & 98.9 & 99.2 & 96.9 & 97.1 & 98.8 & 93.0 & 95.5 \\
\bottomrule
\end{tabular}
\caption{Accuracy (\%) of truthfulness evaluation with the correct and incorrect answers provided by MTruthfulQA dataset.}
\label{tab:evaluator1}
\end{table*}

\begin{table*}[]
\centering
\begin{tabular}{llllllllll}
\toprule
    & \textbf{En} & \textbf{Fr} & \textbf{De} & \textbf{Es} & \textbf{Zh} & \textbf{Ja} & \textbf{Ru} & \textbf{Th} & \textbf{Ar} \\ \hline
\textbf{True Acc. (\%)} & 93.6        & 92.3        & 92.9        & 92.1        & 92.9        & 92.7        & 91.5        & 87.1        & 90.7        \\
\textbf{Info Acc. (\%)} & 99.5        & 98.3        & 98.6        & 98.2        & 98.1        & 97.8        & 98.0        & 92.5        & 96.7 \\
\bottomrule
\end{tabular}
\caption{Accuracy (\%) of truthfulness and informativeness evaluation with responses generated by \textbf{Llama-3-8B+\FMT}.}
\label{tab:evaluator2}
\end{table*}

\end{document}